%% file: main.tex
\documentclass[letterpaper,journal]{IEEEtran}
\usepackage{amsmath,amsfonts}
\usepackage{algorithmic}
\usepackage{algorithm}
\usepackage{array}
\usepackage[caption=false,font=normalsize,labelfont=sf,textfont=sf]{subfig}
\usepackage{textcomp}
\usepackage{stfloats}
\usepackage{url}
\usepackage{verbatim}
\usepackage{graphicx}
\usepackage{cite}
\usepackage{amssymb}
\usepackage{multirow}
\usepackage{multicol}
\usepackage[table]{xcolor}
\usepackage{microtype}
\usepackage{multirow}
\usepackage{amsmath}
\usepackage{amssymb}
\usepackage{graphicx}
\usepackage{algorithmic}
\usepackage{array}
\usepackage{url}
\usepackage{booktabs}
\usepackage{microtype}
\usepackage{float}
\usepackage{bbm}
\usepackage{enumitem}
\usepackage{xcolor}
\usepackage{booktabs}
\usepackage{multirow}
\usepackage{graphicx}
\usepackage{pifont}
\usepackage{colortbl}
\usepackage{xcolor}
\usepackage{csquotes}
\usepackage{colortbl}
\usepackage[
    pagebackref=true,
    breaklinks=true,
    colorlinks=true,
    bookmarks=false
]{hyperref}

\usepackage{pifont}
\usepackage{graphicx}

\usepackage{xcolor}
\usepackage{colortbl}
\usepackage{todonotes}
\usepackage{booktabs}
\usepackage{pifont}

\newcommand{\cmark}{\ding{51}}
\newcommand{\xmark}{\ding{55}}
\begin{document}

\title{Organizing Temporal Evidence in Hyperbolic Space for Video Moment Retrieval}

\author{Mengnan~Liu,
        Sanping~Zhou,~\IEEEmembership{Member,~IEEE,} 
        Kun~Xia,
        Jiaxuan~Lin,
        Gang~Hua,~\IEEEmembership{Fellow,~IEEE,}
        and~Le~Wang,~\IEEEmembership{Senior Member,~IEEE}
       
\thanks{This work was supported in part by the National Key Research and Development Project under Grant 2024YFB4708100, the Fundamental and Interdisciplinary Disciplines Breakthrough Plan of the Ministry of Education of China under Grant JYB2025XDXM504, National Natural Science Foundation of China under Grants 62572384, U24A20325, 62503384, and Key Research and Development Plan of Shaanxi Province under Grant 2024PT-ZCK-80.~({\it Corresponding author: Sanping Zhou, E-mail: spzhou@mail.xjtu.edu.cn.})}
\thanks{M. Liu, S. Zhou, J. Lin and L. Wang are with the State Key Laboratory of Human-Machine Hybrid Augmented Intelligence, Institute of Artificial Intelligence and Robotics, Xi'an Jiaotong University, Xi'an, Shaanxi 710049, China. E-mail: Liumn@stu.xjtu.edu.cn,  spzhou@mail.xjtu.edu.cn, linjiaxuan@stu.xjtu.edu.cn and wangle729@outlook.com. 

\IEEEcompsocthanksitem K. Xia is with the School of Computer Science and Technology, Xi'an Jiaotong University, Xi'an, Shaanxi 710049, China. E-mail: kunxia@xjtu.edu.cn.
\IEEEcompsocthanksitem G. Hua is with Amazon Alexa AI, Bellevue, WA 98004 USA. E-mail: ganghua@gmail.com.
}}

\markboth{Journal of \LaTeX\ Class Files,~Vol.~14, No.~8, August~2021}%
{Shell \MakeLowercase{\textit{et al.}}: A Sample Article Using IEEEtran.cls for IEEE Journals}


\maketitle

\input{tables-and-figures/fig-illustration}

\begin{abstract}

Video moment retrieval aims to localize a query-relevant temporal segment in an untrimmed video. Existing methods typically perform video-text alignment in Euclidean space, where representations are embedded in a flat geometry.
However, accurate moment localization requires organizing query-relevant evidence across multiple grounding states, where target regions, boundary transitions, and contextual distractors exhibit distinct temporal structures and relevance patterns. 
Such Euclidean representations lack an explicit geometry to jointly capture these relationships, motivating us to explore structured temporal evidence organization from a geometric perspective.
Specifically, we propose Hyperbolic Temporal Evidence Organization (HTEO), a geometric framework that organizes query-aware temporal memory in hyperbolic space. HTEO maps temporal representations into a Poincar\'{e} ball and organizes temporal evidence through Radial Evidence Organization (REO) and Orthogonal Angular Refinement (OAR). 
REO adjusts temporal tokens' radial positions to encode the structural differences among temporal evidence, while OAR enhances query-conditioned semantic alignment through orthogonal rotation. 
The organized representation is then projected back to the Euclidean tangent space and injected into the decoder through a lightweight residual memory update, enabling hyperbolic geometric cues to be incorporated into standard Transformer decoders while preserving the original cross-modal memory. Furthermore, we introduce radial structure regularization and angular relevance supervision to explicitly guide temporal evidence organization and query-conditioned semantic alignment, respectively.
Extensive experiments and analyses across multiple benchmarks demonstrate that HTEO improves temporal localization while providing an interpretable geometric organization of temporal evidence.

\end{abstract}


\section{Introduction}
\label{introduction}

\IEEEPARstart{V}{ideo} moment retrieval aims to localize a temporal segment in an untrimmed video according to a natural-language query. It is a fundamental task for video-language understanding and has broad applications in video search~\cite{gabeur2020multi}, human-computer interaction~\cite{liang2023simple}, and multimedia retrieval~\cite{moon2023query}.
Existing methods have achieved remarkable progress in video moment retrieval by improving video-language interaction from different perspectives. Cross-modal attention mechanisms~\cite{zhang2019cross} enable fine-grained interactions between query words and video clips, DETR-style decoders~\cite{sun2024tr} formulate moment localization as a query-based set prediction problem, and contrastive video-text alignment~\cite{xu2021videoclip} further enhances semantic matching between language and visual content. 
These designs have significantly advanced temporal grounding by strengthening cross-modal correspondence modeling. Despite these advances, existing frameworks typically organize video-language representations in Euclidean feature spaces, where representations are embedded in a flat geometry and their relationships are implicitly determined by attention-based interactions or similarity-driven correspondence.

Nevertheless, the underlying geometric structure of these representations remains largely unexplored, limiting their ability to explicitly organize temporal evidence. 
As illustrated in Fig.~\ref{fig:illustration}(a), query-relevant tokens and contextual background tokens may occupy nearby regions in Euclidean memory, making them difficult to distinguish through flat similarity alone. This ambiguity can further produce false response peaks on background segments containing distracting visual cues, as shown in Fig.~\ref{fig:illustration}(b).
These observations suggest that moment retrieval should be viewed not only as a video-text correspondence problem but also as a structured temporal evidence organization problem. A language-guided moment is supported by temporal evidence with different grounding characteristics: target regions provide query-relevant evidence, boundary regions capture temporal transitions during localization, and surrounding background segments introduce contextual distractions. Such heterogeneous temporal evidence is difficult to explicitly organize in flat Euclidean memory, where different evidence types may become entangled within the same representation space.

Motivated by this observation, we revisit video moment retrieval from a geometric perspective of temporal evidence organization. Instead of only asking how to better match video and text features, we ask how temporal evidence should be organized for moment retrieval.
Hyperbolic space provides a natural geometry for this purpose, as its negative curvature and exponentially expanding volume are well suited for modeling structured representations.
In particular, the Poincar\'{e} ball provides two interpretable geometric factors: radius and angle, as shown in Fig.~\ref{fig:illustration}(c). 
The radial component provides a structural cue for organizing temporal evidence according to different grounding characteristics, reflecting differences among target regions, boundary transitions, and contextual background.
In contrast, the angular direction captures query-conditioned semantic relevance, indicating whether a temporal token provides evidence aligned with the query.
This radius-angle interpretation introduces a geometric inductive bias for organizing temporal evidence beyond flat Euclidean matching.

Based on the above insight, we propose Hyperbolic Temporal Evidence Organization (HTEO), a geometric framework designed for structured video moment retrieval. Given query-aware temporal memory, HTEO first projects the memory from the Euclidean tangent space into a Poincar\'{e} ball via the exponential map, enabling non-Euclidean geometric reasoning. In this hyperbolic space, temporal evidence is organized through two complementary modules: Radial Evidence Organization (REO) and Orthogonal Angular Refinement (OAR). 
REO exploits the radial structure of the Poincar\'{e} ball to provide a grounding cue, organizing temporal evidence according to different localization characteristics, including target regions, boundary transitions, and contextual background. OAR employs an orthogonal rotation to refine the angular directions of temporal tokens, enhancing query-conditioned semantic alignment while preserving the radial structure.
After hyperbolic organization, the memory is mapped back to the Euclidean tangent space via the logarithmic map. A lightweight residual memory update is further introduced to inject the organized representation into the decoder, preserving the original cross-modal memory while incorporating structured  geometric cues for moment decoding.

To explicitly guide the two geometric factors, we introduce radial structure regularization and angular relevance supervision. The radial regularization constrains the radial distribution of temporal tokens according to their grounding characteristics, while the angular supervision optimizes the query-conditioned semantic relevance by encouraging target evidence to be aligned with the query direction. Together, these objectives facilitate interpretable hyperbolic evidence organization and complement the standard video moment retrieval losses.

Beyond the proposed framework, we conduct a systematic empirical study of hyperbolic geometry for video moment retrieval. We first investigate the effect of different organization locations by comparing unimodal visual/textual features. We further analyze the individual and joint contributions of the two geometric components through radius-only, angle-only, and combined variants, together with explicit angular relevance and radial structure supervision. 
Furthermore, we provide comprehensive quantitative and qualitative analyses to characterize the behavior of hyperbolic representations. We compare Euclidean and hyperbolic feature distributions, visualize radius-angle temporal trajectories, and analyze the influence of geometric factors under different settings, including background confusion, moment length, curvature, and residual scaling. These studies provide an in-depth examination of how hyperbolic geometry affects temporal representation organization for video moment retrieval.

To summarize, our contributions are as follows:

\begin{itemize}

\item We revisit video moment retrieval from a geometric perspective and formulate it as a temporal evidence organization problem beyond conventional flat video-text matching.

\item We propose Hyperbolic Temporal Evidence Organization (HTEO), a geometric framework that organizes query-aware temporal memory in a Poincar\'{e} ball and leverages radius-angle geometry to model temporal evidence structure and query-conditioned relevance.

\item We design Radial Evidence Organization (REO) and Orthogonal Angular Refinement (OAR), together with a lightweight residual memory update, enabling hyperbolic geometric cues to be integrated into Transformer-based moment decoders while maintaining compatibility with existing architectures.

\item We introduce geometric supervision, including radial structure regularization and  angular relevance supervision, to explicitly guide the learning of interpretable hyperbolic evidence representations.

\item Extensive experiments and systematic analyses on multiple video moment retrieval benchmarks demonstrate the effectiveness and interpretability of HTEO.
\end{itemize}

This paper is organized as follows: Section~\ref{sec:related-work} briefly
reviews related work. We present the technical details of the proposed method in
Section~\ref{sec:methodology}. Experimental results and discussions are
presented in Section~\ref{sec:experiments-and-discussions}. Finally, we conclude
in Section~\ref{sec:conclusion}.

\section{Related Work}
\label{sec:related-work}

We briefly review related work on video moment retrieval and hyperbolic representation learning.

\subsection{Video Moment Retrieval}

Video moment retrieval aims to localize a temporal segment in an untrimmed video according to a natural language query. Early methods formulate this task as cross-modal matching by ranking densely sampled candidate segments according to their semantic similarity with the query~\cite{gao2017tall}. Later works improve efficiency through proposal-free localization, temporal interaction modeling, and span-based boundary prediction~\cite{zhang2021natural}, while a comprehensive survey of temporal sentence grounding is provided in~\cite{zhang2023temporal}.
Beyond temporal-only localization, video grounding has been extended to object-level, spatio-temporal, and pixel-level settings. Representative studies include weakly-supervised video object grounding~\cite{wang2022weakly}, single-frame supervised spatio-temporal grounding~\cite{liu2024single}, and referring video segmentation with language-aware spatial-temporal collaboration and local-global context modeling~\cite{hui2023language,liang2023local}. Recent works further explore structural spatio-temporal alignment~\cite{fei2024enhancing}, efficient temporal filtering for video grounding~\cite{hu2025unified}, and MLLM-based temporal grounding and long-video reasoning~\cite{wu2025survey,li2026momentor++}.

However, these approaches mainly focus on interaction mechanisms and decoding strategies, while the geometric organization of temporal memory remains unexplored.
In contrast, our work views moment retrieval as a temporal evidence organization problem, aiming to structure query-relevant evidence, boundary transitions, and distracting background cues beyond flat Euclidean matching.

\subsection{Hyperbolic Representation Learning}

Hyperbolic space has been widely studied for representing hierarchical and tree-like structures. Compared with Euclidean space, hyperbolic geometry has negative curvature and exponentially increasing volume, making it suitable for embedding data with latent hierarchies~\cite{nickel2017poincare,ganea2018hyperbolic}. Poincar\'{e} embeddings were first introduced to model symbolic hierarchies and taxonomies~\cite{nickel2017poincare}, and later extended to neural networks and graph-structured data~\cite{ganea2018hyperbolic,chami2019hyperbolic}.
In computer vision and vision-language learning, hyperbolic geometry has also been explored for visual hierarchy modeling and cross-modal representation learning. Hyperbolic image embeddings demonstrate the effectiveness of non-Euclidean geometry for visual recognition and semantic hierarchy modeling~\cite{khrulkov2020hyperbolic}, while hyperbolic image-text representations further show that hyperbolic space can improve vision-language alignment by modeling hierarchical semantic structures~\cite{desai2023hyperbolic}. These works often exploit the radial structure of the Poincar\'{e} ball to encode semantic granularity, where abstract concepts are placed near the origin and fine-grained concepts lie farther away. In this geometry, radius naturally reflects hierarchical level or granularity, while angular direction captures semantic similarity or alignment.

Unlike existing hyperbolic vision-language methods that mainly focus on image-level recognition, image-text representation learning, or category-level semantic hierarchies, our work investigates hyperbolic geometry for video moment retrieval.
Although moment retrieval does not contain explicit taxonomic hierarchies, its temporal evidence naturally exhibits structured relationships among target, transition, and contextual regions. We therefore reinterpret radius as a structural cue for temporal evidence granularity and angle as a cue for query-conditioned semantic relevance.

\input{tables-and-figures/fig-framework}

\section{Proposed Method}
\label{sec:methodology}

\subsection{Overview}

Given an untrimmed video and a natural language query, video moment retrieval aims to localize the temporal segment that semantically corresponds to the query. Existing DETR-style methods usually encode video and text features, construct a cross-modal temporal memory, and decode temporal spans from this memory. 
Although such Euclidean memory is effective for video-text interaction, it represents temporal evidence in a flat geometry and lacks an explicit mechanism to organize different grounding characteristics of moment evidence.

We argue that a query-relevant moment is supported by query-relevant temporal evidence at different grounding levels, including target evidence, boundary transitions, and semantically related background distractors. To organize such evidence, we introduce \textbf{Hyperbolic Temporal Evidence Organization (HTEO)}, as illustrated in Fig.~\ref{fig:framework}. Given video and text features, we first construct query-aware temporal memory through cross-modal interaction. HTEO then maps this Euclidean memory into a Poincar\'{e} ball and organizes temporal evidence through Radial Evidence Organization (REO) and Orthogonal Angular Refinement (OAR). The organized representation is mapped back to the Euclidean tangent space and injected into the moment decoder through a lightweight residual update, enabling compatibility with standard decoders while preserving the original cross-modal memory.


\input{tables-and-figures/angle_speration}
\subsection{Feature Extraction}

\noindent\textbf{Visual Features.} Following prior work~\cite{lei2021detecting}, the video is first divided into non-overlapping clips according to a certain time interval. Then the pre-trained video encoder is utilized to extract clip-level visual features $\mathbf{F}_{v} = [f_{v}^{1}, f_{v}^{2},..., f_{v}^{N_v}] \in \mathbb{R}^{N_v \times d_{v}}$, where $N_v$ and $d_v$ are the number of clips and the visual feature dimension, respectively. 
\newline
\textbf{Textual Features.} For a natural language query, we use the pre-trained textual encoder to extract textual features $\mathbf{F}_{t} \in \mathbb{R}^{N_t \times d_{t}}$ following~\cite{zhao2025ld}, where $N_t$ and $d_{t}$ indicate the number of words and the textual feature dimension, respectively.

\subsection{Geometric Analysis of Temporal Evidence Organization}
Moment retrieval requires organizing structured temporal evidence rather than simply matching a query to video features. A query-relevant moment is not determined by a single correspondence, but is supported by heterogeneous temporal evidence with different grounding characteristics, including target evidence, boundary transitions, and semantically related background contexts. Therefore, temporal memory should not only measure semantic similarity, but also preserve the structural relationships among different evidence states.

\subsubsection{Limitations of Euclidean Geometry}
From a geometric perspective, conventional temporal memory is
represented in a flat Euclidean space, where the relationships
between tokens are mainly characterized by pairwise
distances,
\begin{equation}
d_{\mathbb{E}}(\mathbf{x},\mathbf{y})
=
\|\mathbf{x}-\mathbf{y}\|_2.
\end{equation}
Euclidean representations can also be decomposed into magnitude
and direction. However, this decomposition does not introduce additional geometric capacity for organizing heterogeneous temporal evidence. More specifically, the Euclidean metric in hyperspherical
coordinates is given by~\cite{cannon1997hyperbolic},
\begin{equation}
ds_{\mathbb{E}}^2
=
dr^2+r^2d\Omega^2,
\end{equation}
where $r$ denotes the radial distance and $d\Omega$ represents an
infinitesimal angular displacement. For two tokens located at a
similar radius, the contribution of angular displacement increases quadratically with the radius through the term $r^2d\Omega^2$. Moreover, the volume of a
$d$-dimensional Euclidean ball grows only polynomially,
\begin{equation}
V_{\mathbb{E}}(r)\propto r^d.
\end{equation}

Although such a flat geometric structure does not prevent Euclidean representations from modeling temporal evidence, different evidence states still need to be organized within the same shared embedding space under a fixed dimensionality. Due to the zero-curvature and homogeneous metric of Euclidean geometry, distances and directions mainly reflect local similarity relationships without introducing an explicit geometric bias for structural organization. Consequently, when temporal tokens with similar visual semantics but different grounding roles are embedded into nearby regions, flat similarity matching may provide insufficient discrimination between target evidence and contextual distractors.

\subsubsection{Why Hyperbolic Geometry}
A hyperbolic space with constant sectional curvature $-c$, where
$c>0$, provides a fundamentally different radius--angle geometry
from its Euclidean counterpart~\cite{cannon1997hyperbolic,krioukov2010hyperbolic}. In geodesic
polar coordinates, 
\begin{equation}
ds_{\mathbb{H}}^2
=
dr^2
+
\left(
\frac{\sinh(\sqrt{c}r)}{\sqrt{c}}
\right)^2
d\Omega^2.
\end{equation}
Since
\begin{equation}
\frac{\sinh(\sqrt{c}r)}{\sqrt{c}}>r,
\qquad r>0,
\end{equation}
the same angular deviation produces a larger geodesic separation
in hyperbolic space than in Euclidean space, particularly for
representations located farther from the origin. In addition, the
volume of a hyperbolic ball grows approximately exponentially,
\begin{equation}
V_{\mathbb{H}}(r)
\propto
e^{(d-1)\sqrt{c}r},
\end{equation}
which contrasts with the polynomial volume growth of Euclidean
space and provides increasingly greater representational capacity
as the radius grows~\cite{nickel2017poincare,krioukov2010hyperbolic}.

The interaction between radius and angle can be further characterized
by the hyperbolic law of cosines. For two hyperbolic tokens
$\mathbf{h}_i$ and $\mathbf{h}_j$ with radial distances $r_i$ and
$r_j$ and an angular difference $\theta_{ij}$, their geodesic
distance satisfies,
\begin{equation}
\begin{split}
\cosh\left(
\sqrt{c}\,
d_{\mathbb{H}}(\mathbf{h}_i,\mathbf{h}_j)
\right)
={}&
\cosh(\sqrt{c}r_i)\cosh(\sqrt{c}r_j)
\\
&-
\sinh(\sqrt{c}r_i)\sinh(\sqrt{c}r_j)
\cos\theta_{ij}.
\end{split}
\end{equation}
Unlike the linear radius--angle interaction in Euclidean geometry,
this relation shows that radial position and angular deviation are
nonlinearly coupled by negative curvature. Therefore, even a modest
angular difference can induce a substantially larger geodesic
separation when the corresponding evidence is located at a larger
radius.

This geometric property is well aligned with temporal evidence organization in moment retrieval. Specifically, the radial component captures structural variations among different temporal evidence states, while the angular component improves the discrimination between target evidence and semantically similar distractors. To empirically examine this geometric interpretation, we visualize and analyze the radius-angle characteristics of hyperbolic representations in Fig.~\ref{fig:intro_radius_angle}.

As shown in Fig.~\ref{fig:intro_radius_angle}(a), we visualize the hyperbolic radius and its time gradient for a representative query, where the shaded region denotes the ground-truth moment and dashed lines indicate its boundaries. The radius shows clear changes near the start and end boundaries, and the gradient response peaks around boundary regions. This observation suggests that the radial coordinate captures temporal structural variations and is sensitive to boundary transitions, providing an additional geometric cue for temporal evidence organization.

We further compare the per-query hyperbolic angular similarity among GT, boundary, and background tokens in Fig.~\ref{fig:intro_radius_angle}(b). GT tokens generally exhibit higher angular similarity than boundary and background tokens, while boundary tokens tend to lie between target and non-relevant regions. This pattern suggests that the angular component provides a relevance-aware cue for discriminating temporal evidence in hyperbolic space.
Accordingly, we adopt the following radius-angle interpretation:
\begin{equation}
\text{radius}
\Rightarrow
\text{temporal evidence structure},
\end{equation}
and
\begin{equation}
\text{angle}
\Rightarrow
\text{query-conditioned semantic relevance}.
\end{equation}

These two factors jointly define the geodesic organization of temporal evidence in hyperbolic space, providing HTEO with a geometry-aware inductive bias.

\subsection{Hyperbolic Temporal Evidence Organization}

Given the query-aware temporal memory, HTEO first maps the Euclidean representation into a Poincar\'{e} ball through the exponential map, enabling radius-angle based geometric organization. Based on this task-specific geometric interpretation, HTEO introduces two complementary components: Radial Evidence Organization (REO), which adjusts the structural positions of temporal tokens to capture different grounding states and temporal transitions, and Orthogonal Angular Refinement (OAR), which refines semantic directions to improve query-conditioned discrimination. Together, the radial and angular components determine the geodesic organization of temporal evidence under hyperbolic geometry.


\paragraph{Exponential Map}
The temporal memory $\mathbf{M}$ resides in the Euclidean tangent space, where semantic evidence is implicitly mixed and difficult to explicitly organize. To introduce an inductive bias for temporal modeling and query-conditioned relevance, we map $\mathbf{M}$ into a Poincar\'{e} ball using the exponential map:
\begin{equation}
    \mathbf{H} = \exp_0^c(\mathbf{M}),
\end{equation}
where \(c>0\) denotes the curvature magnitude, and the
sectional curvature of the Poincar\'{e} ball is \(K=-c\). $\mathbf{H} = \{\mathbf{h}_t\}_{t=1}^{T}$ denotes the hyperbolic memory. 
This mapping enables each temporal token to be characterized by radial and angular coordinates in the hyperbolic space.

\paragraph{Radial Evidence Organization}

After mapping the memory into the Poincar\'{e} ball, each hyperbolic token is associated with a radial coordinate, measured by its geodesic distance to the origin,
\begin{equation}
    r_t = d_{\mathbb{H}}^c(0, \mathbf{h}_t).
\end{equation}

To adapt the radial organization of temporal evidence, we introduce a radius-aware transformation based on Möbius matrix–vector multiplication. Given the hyperbolic temporal tokens $\mathbf{H}={\{\mathbf{h}_t}\}_{t=1}^{T}$, the transformation is applied as follows,
\begin{equation}
\mathcal{S}(\mathbf{h}_t)
=
\mathbf{S}\otimes_c \mathbf{h}_t,
\end{equation}
where $\mathbf{S}$ is a learnable block-diagonal transformation defined over the feature dimensions, and $\otimes_c$ denotes Möbius matrix–vector multiplication in a hyperbolic space with curvature c. By adjusting the hyperbolic magnitude and orientation of individual temporal tokens while preserving the underlying manifold geometry, this operation reorganizes temporal evidence and provides structured radial cues for temporal grounding.

Since geometric refinement may introduce undesirable perturbations to the original cross-modal memory, we introduce a geometry-consistent residual update that balances radial adaptation with representation preservation. Specifically, for each temporal token $\mathbf{h}_t$, the refined representation is computed as follows,
\begin{equation}
\bar{\mathbf{h}}_t
=
\exp_{\mathbf{h}_t}^{c}
\left(
\lambda_r
\log_{\mathbf{h}_t}^{c}
\big(
\mathcal{S}(\mathbf{h}_t)
\big)
\right),
\end{equation}
where $\log_{\mathbf{h}_t}^{c}(\cdot)$ maps the geometrically transformed token to the tangent space at $\mathbf{h}_t$, and $\exp_{\mathbf{h}_t}^{c}(\cdot)$ maps the scaled displacement back to the Poincar\'{e} ball. The coefficient $\lambda_r\in[0,1]$ controls the degree of geometric refinement: $\lambda_r=0$ preserves the original memory, whereas $\lambda_r=1$ fully adopts the transformed representation. This update interpolates between the original and refined tokens along the underlying hyperbolic geodesic, thereby mitigating excessive geometric distortion while preserving manifold consistency.

\paragraph{Orthogonal Angular Refinement}
Although radial coordinates capture structural variations among temporal evidence, they cannot distinguish whether a token is semantically aligned with the query. Temporal tokens with similar radial characteristics may still contain different semantic relevance. Therefore, after radial organization, we introduce angular refinement to adjust the semantic orientation of hyperbolic tokens and enhance query-conditioned relevance discrimination.

For each radius-aware hyperbolic token $\bar{\mathbf{h}}_t$, we first map it to the tangent space at the origin,
\begin{equation}
\mathbf{u}_t
=
\log_{0}^{c}(\bar{\mathbf{h}}_t).
\end{equation}
Let $\mathbf{u}_q \in \mathbb{R}^{d}$ denote the global query representation projected into the same tangent space. Since the Poincar\'{e} ball is conformal, the angular relation between the temporal token and the query can be measured by their tangent-space cosine similarity,
\begin{equation}
s_t
=
\frac{
\mathbf{u}_t^{\top}\mathbf{u}_q
}{
\|\mathbf{u}_t\|_2
\|\mathbf{u}_q\|_2
}.
\end{equation}
We then introduce a lightweight query-conditioned angular gate,
\begin{equation}
g_t
=
\sigma\left(
\beta(1-s_t)+b
\right),
\end{equation}
where $\beta$ and $b$ are learnable parameters. The gate $g_t$ controls the rotation magnitude according to the angular discrepancy between the $t$-th temporal token and the query. Tokens already aligned with the query require weaker angular adjustment, whereas less relevant tokens receive stronger directional refinement.

Let $\mathbf{R}_t \in \mathbb{R}^{d \times d}$ be an orthogonal rotation matrix satisfying
$\mathbf{R}_t^{\top}\mathbf{R}_t=\mathbf{I}$. For each tangent-space token $\mathbf{u}_t$, we apply the angular refinement and map it back to the Poincar\'{e} ball as follows,
\begin{equation}
    \widetilde{\mathbf{h}}_t = \exp_{0}^{c}
\left(
\mathbf{R}_t\mathbf{u}_t
\right),
\end{equation}
 and then obtain the hyperbolic memory $\widetilde{\mathbf{H}} =\{\widetilde{\mathbf{h}}_t\}_{t=1}^{T}$.

Since $\mathbf{R}_t$ is an orthogonal matrix, it preserves the Euclidean norm in the tangent space,
\begin{equation}
\|\mathbf{R}_t\mathbf{u}_t\|_2
=
\|\mathbf{u}_t\|_2.
\end{equation}
Under the exponential map, the hyperbolic geodesic distance from the origin is determined by the tangent-space norm, and thus the radial coordinate is preserved. Accordingly, we have,
\begin{equation}
d_{\mathbb{H}}^c
(0,\widetilde{\mathbf{h}}_t)
=
d_{\mathbb{H}}^c
(0,\bar{\mathbf{h}}_t).
\end{equation}
Thus, this operation adjusts the query-relative angular direction of temporal tokens while preserving their radial structure. This property is crucial for our radius--angle decomposition, as it enables orthogonal angular refinement to refine query-conditioned semantic relevance without disrupting the temporal structure encoded in the radial component.

In implementation, we parameterize $\mathbf{R}_t$ using the Cayley transform to guarantee orthogonality during optimization. Given a learnable matrix
$\Theta \in \mathbb{R}^{d \times d}$, we first construct a token-specific skew-symmetric matrix,
\begin{equation}
\mathbf{A}_t
=
g_t
\left(
\Theta-\Theta^{\top}
\right).
\end{equation}
The rotation matrix is then obtained as follows,
\begin{equation}
\mathbf{R}_t
=
(\mathbf{I}-\mathbf{A}_t)^{-1}
(\mathbf{I}+\mathbf{A}_t).
\end{equation}
This parameterization guarantees
$\mathbf{R}_t^{\top}\mathbf{R}_t=\mathbf{I}$, enabling stable
angular semantic refinement while preserving the radial structure.
Compared with unconstrained linear transformations, orthogonal rotations
avoid arbitrary feature scaling and better preserve the disentangled
radius--angle representation in hyperbolic memory.

\paragraph{Logarithmic Map}
After REO and OAR, the refined memory lies in the Poincar\'{e} ball. However, the subsequent Transformer decoder and prediction heads operate in Euclidean feature space. To keep the original decoder architecture unchanged, we map the hyperbolic memory back to the Euclidean tangent space through the logarithmic map,
\begin{equation}
\widetilde{\mathbf{M}}
=
\log_0^c(\widetilde{\mathbf{H}}),
\end{equation}
where $\widetilde{\mathbf{M}}\in\mathbb{R}^{T\times d}$ denotes the Euclidean representation obtained after hyperbolic organization. In this design, hyperbolic geometry serves as a memory organization mechanism: temporal tokens are first organized in a non-Euclidean space to introduce radius-angle geometric structures for evidence relevance and temporal organization, and the resulting representations are then projected back to the Euclidean tangent space for standard Transformer decoding.


\subsection{Residual Memory Update}

To retain the learned cross-modal representation while benefiting from hyperbolic organization, we incorporate the refined memory through a lightweight residual update. This design preserves the integrity of the learned video-text embeddings while injecting structured  geometric information for improved temporal reasoning.

We define the hyperbolic organization direction as $\Delta \mathbf{M} = \widetilde{\mathbf{M}} - \mathbf{M}$. The decoder memory is obtained through a residual update,
\begin{equation}
    \mathbf{M}_{\mathrm{d}} = \mathbf{M} + \alpha(\widetilde{\mathbf{M}} - \mathbf{M}),
\end{equation}
where $\alpha$ is a residual scaling coefficient that controls how strongly the hyperbolic refinement direction is injected into the original Euclidean memory.


\subsection{Geometric Regularization}

To explicitly regularize the hyperbolic evidence organization, we introduce two complementary geometric losses that constrain the radial structure and angular relevance of the learned representations. Specifically, the radial structure loss encourages meaningful temporal evidence organization, while the angular relevance loss enhances query-conditioned semantic alignment.


\paragraph{Radial Structure Loss}

The radial coordinate provides a structural cue for organizing temporal evidence with different grounding characteristics. We therefore define a radial constraint directly in the Poincar\'{e} ball based on the geodesic distance to the origin. For each hyperbolic token $\widetilde{\mathbf{h}}_t$, its radius is computed as $r_t = d_{\mathbb{H}}^c (0, \widetilde{\mathbf{h}}_t).$
To encourage radius to encode discriminative temporal evidence structure, we compute the average radii of ground-truth moment tokens and background tokens,
\begin{equation}
r_{\mathrm{gt}}
=
\frac{1}{|\mathcal{P}|}
\sum_{t\in\mathcal{P}} r_t,
\quad
r_{\mathrm{bg}}
=
\frac{1}{|\mathcal{N}|}
\sum_{t\in\mathcal{N}} r_t.
\end{equation}

Instead of imposing a fixed ordering between moment and background radii, we only encourage their radial distributions to be separable,
\begin{equation}
\mathcal{L}_{\mathrm{radius}}^{\mathbb{H}}
=
\max
\left(
0,
\gamma_{\mathrm{rad}}
-
\left|
r_{\mathrm{gt}}
-
r_{\mathrm{bg}}
\right|
\right),
\end{equation}
where $\gamma_{\mathrm{rad}}$ is a radial margin, $\mathcal{P}$ and $\mathcal{N}$ denote the sets of temporal tokens inside and outside the ground-truth moment, respectively. This formulation avoids assuming that moment tokens should always lie closer to or farther from the origin than background tokens, since radial distributions may vary across different videos and datasets. Instead, it encourages the radial coordinate to capture discriminative temporal evidence structures.

Since the radial coordinate is expected to explicitly characterize temporal transitions, we introduce a transition-aware radial regularizer. Let $\mathcal{B}$ denote the set of adjacent temporal pairs located within a small neighborhood of the ground-truth start and end boundaries, and let $\mathcal{I}$ denote the set of non-boundary temporal pairs. We first define the radial variation between two adjacent temporal positions as $\Delta_t = \left|r_{t+1}-r_t\right|$. We then encourage the radial variations around temporal boundaries to be larger than those in non-boundary regions,
\begin{equation}
\mathcal{L}_{\mathrm{trans}}^{\mathbb{H}}
=
\max
\left(
0,
\frac{1}{|\mathcal{I}|}
\sum_{t\in\mathcal{I}}
\Delta_t
-
\frac{1}{|\mathcal{B}|}
\sum_{t\in\mathcal{B}}
\Delta_t
\right).
\end{equation}

The final radial regularization objective is formulated as follows,
\begin{equation}
\mathcal{L}_{\mathrm{rad}}^{\mathbb{H}}
=
\mathcal{L}_{\mathrm{radius}}^{\mathbb{H}}
+
\lambda_{\mathrm{trans}}
\mathcal{L}_{\mathrm{trans}}^{\mathbb{H}},
\end{equation}
where $\lambda_{\mathrm{trans}}$ controls the contribution of the transition-aware constraint. This objective explicitly enhances radial changes around target--context transitions, making the learned radial geometry more sensitive to temporal boundaries.


\paragraph{Angular Relevance Loss}

While the radial coordinate captures structural variations among temporal evidence, it does not explicitly measure whether a token is relevant to the given query. Therefore, we introduce an angular relevance constraint to encourage query-conditioned semantic alignment.

For a hyperbolic token $\widetilde{\mathbf{h}}_t$, we first aggregate GT tokens in the tangent space and then map the resulting semantic direction back to the hyperbolic space as follows,

\begin{equation}
\mathbf{m}_g =
\frac{1}{|\mathcal{P}|}
\sum_{t \in \mathcal{P}}
{\log}_0^c(\widetilde{\mathbf{h}}_t), \quad
\mathbf{h}_g = {\exp}_0^c(\mathbf{m}_g).
\end{equation}

The relevance score of each temporal token is then computed by measuring the geodesic angular similarity between the token direction and the ground-truth evidence direction in the Poincar\'{e} ball,

\begin{equation}
s_t =
\operatorname{sim}_{\mathbb H}^c
(\widetilde{\mathbf{h}}_t,\mathbf{h}_g)
=
\cos
\left(
\theta_{\mathbb H}^c
(\widetilde{\mathbf{h}}_t,\mathbf{h}_g)
\right),
\end{equation}
where $\theta_{\mathbb H}(\cdot,\cdot)$ denotes the geodesic angle between two directions in the Poincar\'{e} ball, and the cosine transformation converts the angular deviation into a relevance score.

We then compute the positive and negative relevance scores as follows,
\begin{equation}
s_{\mathrm{pos}}
=
\frac{1}{|\mathcal{P}|}
\sum_{t\in\mathcal{P}} s_t,
\quad
s_{\mathrm{neg}}
=
\frac{1}{|\mathcal{N}|}
\sum_{t\in\mathcal{N}} s_t.
\end{equation}
The hyperbolic angular relevance loss is then defined as follows,
\begin{equation}
\mathcal{L}_{\mathrm{ang}}^{\mathbb{H}}
=
\max
\left(
0,
\gamma_{\mathrm{ang}}
-
s_{\mathrm{pos}}
+
s_{\mathrm{neg}}
\right),
\end{equation}

where $\gamma_{\mathrm{ang}}$ is a margin. This loss encourages tokens within the ground-truth moment to exhibit smaller angular deviations from the target direction than background tokens, thereby explicitly regularizing the angular component to distinguish target-relevant temporal evidence from background context.

\subsection{Training Objective}
The final training objective is written as follows,
\begin{equation}
\mathcal{L}_{\mathrm{total}}
=
\mathcal{L}_{\mathrm{MR}}
+
\lambda_{\mathrm{ang}}
\mathcal{L}_{\mathrm{ang}}^{\mathbb{H}}
+
\lambda_{\mathrm{rad}}
\mathcal{L}_{\mathrm{rad}}^{\mathbb{H}},
\end{equation}
where $\mathcal{L}_{\mathrm{MR}}$ denotes the standard moment retrieval loss following LD-DETR~\cite{zhao2025ld}, while $\lambda_{\mathrm{ang}}$ and $\lambda_{\mathrm{rad}}$ control the strengths of the geometric regularizers.

\section{Experiments and Discussion}
\label{sec:experiments-and-discussions}

In this section, we first introduce three standard benchmarks, the evaluation metrics, and the implementation details. Then, we compare the proposed HTEO with state-of-the-art methods, followed by a set of ablation studies.

\input{tables-and-figures/tab-comparison-on-thumos}

\input{tables-and-figures/tab-comparison-on-activitynet13}

\subsection{Datasets}

\textbf{QVHighlights}~\cite{lei2021detecting} is a widely used benchmark for both video grounding and highlight detection. It contains 10,148 videos, 18,367 annotated moments, and 10,310 natural language sentences. Each moment is labeled with a five-level highlight score ranging from \enquote{very poor} to \enquote{excellent}, enabling joint evaluation of moment localization and highlight detection. Since the annotations of the test set are withheld, we follow the standard protocol and report results on the official validation set.

\textbf{Charades-STA}~\cite{gao2017tall} is a standard benchmark for temporal moment retrieval. It consists of 9,848 indoor activity videos and 16,128 natural language queries, each aligned with a specific temporal segment. The videos are approximately 30 seconds long on average, while the annotated moments have an average duration of about 8.1 seconds. Following the official split, we use 12,408 moment-query pairs for training and 3,720 pairs for testing, ensuring consistent and reproducible evaluation.

\textbf{TACoS}~\cite{regneri2013grounding} is a fine-grained video-language grounding benchmark built from high-resolution cooking videos. It contains 17,334 action descriptions, comprising 146,771 words and 75,210 lexical instances. The videos range from 1 to 23 minutes in duration, with an average length of approximately 4.5 minutes, providing diverse temporal spans for fine-grained video-language alignment. Following the official split, we use 10,146 training pairs, 4,589 validation pairs, and 4,083 testing pairs for evaluation.

\subsection{Metrics and Implementation Details}
\textbf{Metrics.}
Following recent studies such as Moment-DETR~\cite{lei2021detecting}, QD-DETR~\cite{moon2023query}, and TR-DETR~\cite{sun2024tr}, we adopt standard evaluation metrics for video moment retrieval and highlight detection. For QVHighlights, we report Recall@1 under IoU thresholds of 0.5 and 0.7, denoted as R1@0.5 and R1@0.7. We also report mean average precision (mAP) at IoU thresholds of 0.5 and 0.75. Following Moment-DETR~\cite{lei2021detecting}, we further compute the average mAP over 10 evenly sampled IoU thresholds from 0.5 to 0.95. For highlight detection, we use mAP and HIT@1 as the primary evaluation metrics. For Charades-STA and TACoS, we evaluate Recall@1 with IoU thresholds of 0.3, 0.5, and 0.7.

\textbf{Implementation Details.}
For all experiments, the hidden dimension $d_{\mathrm{model}}$ is set to 256. On QVHighlights, we use SlowFast~\cite{feichtenhofer2019slowfast} and CLIP visual features, and extract textual features using the CLIP text encoder. The model is trained for 200 epochs with a batch size of 32 and a learning rate of $1\times10^{-4}$.
For Charades-STA, we use visual features extracted from VGG~\cite{sengupta2019going}, I3D~\cite{carreira2017quo}, SlowFast~\cite{feichtenhofer2019slowfast}, and CLIP. Textual features are represented by GloVe embeddings~\cite{pennington2014glove}. The model is trained for 100 epochs with a batch size of 8 and a learning rate of $1\times10^{-4}$.
For TACoS, we use C3D~\cite{tran2015learning} visual features and GloVe~\cite{pennington2014glove} textual features. The training process lasts for 200 epochs, with a batch size of 32 and a learning rate of $2\times10^{-4}$.
All experiments are conducted on a single NVIDIA RTX 3090 GPU.

\subsection{Comparison with Other Methods}

Table~\ref{tab:qvhighligts} reports the comparison on the QVHighlights dataset. HTEO consistently improves the performance of different Transformer-based moment retrieval frameworks, demonstrating its effectiveness as a general geometric evidence organization module. When integrated into TR-DETR, HTEO improves the Avg. mAP from 46.42 to 47.63. More importantly, applying HTEO to LD-DETR further boosts the Avg. mAP from 47.18 to 48.42 and improves HIT@1 from 66.39 to 68.26, achieving the best performance on both moment retrieval and highlight detection. The improvements under stricter localization metrics, including R1@0.7 and mAP@0.75, indicate that HTEO provides more precise temporal evidence organization for accurate moment boundary localization.

We further evaluate HTEO on TACoS and Charades-STA, as shown in Table~\ref{tab:tacos and charades}. On TACoS, HTEO consistently improves LD-DETR across all evaluation metrics, increasing R1@0.7 from 26.2 to 27.0 and mIoU from 40.3 to 41.5, demonstrating improved capability in fine-grained temporal localization. Similarly, on Charades-STA, HTEO achieves consistent gains over the strong LD-DETR baseline, improving R1@0.7 from 41.6 to 42.9 and mIoU from 53.4 to 54.1. These results demonstrate that the proposed hyperbolic evidence organization framework generalizes well across datasets with different video characteristics.

Overall, the consistent improvements across multiple architectures, datasets, and evaluation metrics validate the effectiveness of HTEO in enhancing temporal evidence organization for video moment retrieval.

\input{tables-and-figures/fig-hyperbolic}


\subsection{Ablation Analysis of Angle and Radius}
We further investigate the contributions of the radial and angular components in HTEO, as shown in Table~\ref{tab:radius-angle calibration}. Starting from the Euclidean baseline, we evaluate each geometric component individually and their combination.
Applying radial organization alone improves the overall localization performance, indicating that the radial component provides useful structural cues for organizing temporal evidence. However, its limited gain compared with angular modeling suggests that temporal structure alone is insufficient without query-conditioned relevance.
In contrast, angular organization consistently improves the retrieval performance, demonstrating its effectiveness in capturing query-aware semantic relevance. By jointly modeling these two complementary components, HTEO achieves the best overall performance, further validating their synergy in structural modeling and semantic alignment and yielding a more structured hyperbolic representation for video moment retrieval.
\input{tables-and-figures/tab-ablation-radius-angle}

\subsection{Effect of Explicit Geometric Supervision}
We further investigate the effect of geometric supervision in Table~\ref{tab:geometric losses}. Starting from the HTEO model without additional geometric constraints, we evaluate the individual and joint effects of radial and angular supervision with different weighting configurations.
Introducing radial supervision consistently improves localization performance, indicating that explicit radial constraints help organize temporal evidence with different grounding characteristics. Angular supervision also provides complementary benefits by enhancing query-conditioned relevance modeling, particularly improving the retrieval accuracy under stricter localization metrics. However, optimizing either geometric factor alone provides limited gains, suggesting that temporal evidence organization requires both structural and semantic constraints.
By jointly applying radial and angular supervision, HTEO achieves the best overall performance. These results demonstrate that the two geometric constraints play complementary roles and provide effective guidance for learning structured hyperbolic representations for moment retrieval.
\input{tables-and-figures/tab-ablation-loss}

\subsection{Analysis under Temporal Evidence Ambiguity}

We further investigate the robustness of HTEO under different levels of temporal evidence ambiguity, as shown in Table~\ref{tab:Background confusion analysis}. Samples are grouped into low, medium, and high ambiguity levels according to the background confusion scores, and the performance of the baseline and HTEO is compared under each setting.
HTEO consistently outperforms the baseline across different ambiguity levels. Under low ambiguity, HTEO achieves clear improvements, indicating that hyperbolic evidence organization effectively enhances the separation between target evidence and contextual distractions when the temporal evidence is relatively distinguishable. For medium ambiguity cases, HTEO maintains comparable performance with the baseline, demonstrating stable behavior under moderate evidence uncertainty.
More importantly, under high ambiguity scenarios, where background segments introduce stronger interference and are more difficult to distinguish from the target moment, HTEO still achieves consistent improvements over the baseline. This suggests that the proposed hyperbolic representation provides a more structured organization of temporal evidence and improves the robustness of moment localization beyond flat similarity-based matching.

Overall, these results demonstrate that HTEO is effective across varying levels of temporal evidence ambiguity, validating the benefit of geometric evidence organization for robust video moment retrieval.
\input{tables-and-figures/tab-background-confusion}

\subsection{Moment Length Analysis}
We further investigate the effect of temporal moment length on HTEO to understand its behavior across different temporal scales. As shown in Table~\ref{tab:momoent length}, ground-truth moments are categorized into short, medium, and long durations, and HTEO is compared with the baseline under identical settings.
HTEO consistently improves the baseline across all duration categories, demonstrating its robustness to different temporal scales. Notably, short-duration moments obtain the largest improvement, indicating that hyperbolic evidence organization is effective in distinguishing highly localized temporal evidence where precise boundary and semantic discrimination are required. For medium- and long-duration moments, HTEO also provides consistent gains, suggesting that the proposed geometric organization can enhance temporal reasoning without being limited by moment length.
Overall, these results demonstrate that HTEO generalizes across different temporal scales and provides a more structured representation for organizing query-relevant temporal evidence.
\input{tables-and-figures/tab-moment-length}

\subsection{Analysis of Target-Background Response Discrimination}

To further quantify the ability of hyperbolic angular organization to distinguish target moments from distracting regions, we analyze the inside-outside response difference across representative queries. As shown in Fig.~\ref{fig:false_peak}, the left panel compares the maximum responses within and beyond the ground-truth interval. Euclidean similarity may assign higher peak responses to outside regions, indicating vulnerability to distractor-induced activations. In contrast, hyperbolic angular similarity increases the response dominance of target moments.
The right panel summarizes the corresponding inside-outside response gaps, where positive values indicate stronger target-region responses. Compared with Euclidean similarity, hyperbolic angular organization shifts the gaps toward positive values, demonstrating improved discrimination between relevant temporal evidence and background distractions.

\input{tables-and-figures/peak}

\input{tables-and-figures/fig-angle}

\subsection{Analysis of Radial Temporal Evidence Structure}

We further examine the radial statistics across different temporal regions to determine whether the hyperbolic radius captures meaningful temporal evidence structure. As shown in Table~\ref{tab:radius_structural_analysis}, the radial distributions exhibit systematic differences among the GT, boundary, and background regions. Specifically, GT tokens have the smallest mean radius and the lowest variance, indicating that query-relevant evidence forms a relatively compact and stable radial distribution. Boundary tokens exhibit a slightly larger mean radius and greater variability, consistent with their transitional role between relevant and irrelevant content. In contrast, background tokens have the largest mean radius and are temporally farther from the center of the GT moment, suggesting a clearer separation from the target evidence.
The close agreement between the mean and median within each region indicates that these radial patterns are stable rather than driven by a few extreme samples. The values in Table~\ref{tab:radius_structural_analysis} are computed from the original, unscaled representations, whereas the qualitative curves amplify the adapter displacement by a factor of five for clearer visualization. Therefore, the larger values in the plots are not directly comparable to those in the table and are intended only to illustrate relative temporal variations in radius.

\input{tables-and-figures/tab-radius_structure}

\subsection{Effect of Residual Memory Update}

To investigate whether the improvements mainly come from hyperbolic organization rather than generic feature adaptation, we compare HTEO with parameter-matched Euclidean adapter controls in Table~\ref{tab:euclidean_adapter_control}. Although Euclidean adapters can slightly improve the baseline performance, their gains remain limited compared with HTEO. This indicates that simply transforming or refining features in Euclidean space is insufficient to capture the structured temporal evidence required for moment localization.
In contrast, HTEO achieves consistent improvements by introducing hyperbolic geometric organization and residual memory refinement. The residual update preserves the original cross-modal memory while incorporating the geometric evidence structure learned in hyperbolic space, enabling effective refinement without disrupting existing multimodal representations. These results demonstrate that the performance gains come from geometry-aware evidence organization rather than generic feature adaptation.
 \input{tables-and-figures/tab-ablation-control}

\subsection{Sensitivity to the Curvature Magnitude $c$}
We further analyze the sensitivity of the hyperbolic curvature parameter $c$ in the Poincar\'{e} ball, where the underlying sectional curvature is defined as $K=-c$. As shown in Table~\ref{tab:Curvature sensitivity}, HTEO achieves the best performance with $c=0.01$, indicating that an appropriate negative curvature magnitude provides a suitable geometric space for temporal evidence organization. When $c$ is smaller, the hyperbolic space becomes closer to Euclidean geometry and provides weaker non-Euclidean geometric separation. In contrast, excessively large curvature magnitudes may introduce overly expanded geometry and reduce representation stability. These results demonstrate the importance of selecting an appropriate curvature magnitude to balance geometric expressiveness and stable temporal evidence organization.
\input{tables-and-figures/tab-ablation-curvaturer}

\input{tables-and-figures/token_distribution}

 \subsection{Ablation on Hyperbolic Injection Position}
To investigate where hyperbolic organization is most effective, we evaluate different injection positions, including unimodal features, pre-fusion representations, and post-fusion query-aware memory, as shown in Table~\ref{tab:hyperbolic calibration}. Applying HTEO to unimodal features provides limited improvements, indicating that individual modalities lack sufficient query-conditioned temporal evidence for effective geometric organization. The pre-fusion setting achieves moderate gains, while applying HTEO after cross-modal fusion obtains the best performance. This suggests that fused query-aware memory provides a more suitable representation space, where temporal evidence structure and query relevance are jointly encoded. Therefore, HTEO is most effective when applied after multimodal interaction, enabling geometric organization of temporal evidence while preserving cross-modal semantics.
\input{tables-and-figures/tab-position-cal}

\subsection{Sensitivity Analysis of the Residual Scaling Factor}
We further analyze the sensitivity of the residual scaling factor $\alpha$ in Table~\ref{tab:Delta scale alpha sensitivity}. 
The best performance is achieved at $\alpha=0.10$, which consistently improves R1@0.5, R1@0.7, and Avg. mAP. When $\alpha$ becomes too small or too large, performance slightly degrades, suggesting that an appropriate balance is required to effectively inject hyperbolic refinement without over-perturbing the original cross-modal representation.
Overall, these results demonstrate that the proposed method is robust to residual scaling and performs stably across different $\alpha$ values.
\input{tables-and-figures/tab-ablation-residual}

\input{tables-and-figures/fig-qualitative}

\subsection{Visualization of Hyperbolic Evidence Organization}

We visualize the learned hyperbolic memory to investigate whether the proposed radius-angle geometry provides an interpretable organization of temporal evidence. As shown in Fig.~\ref{fig:radius_angle}, query-relevant tokens, boundary tokens, and background tokens exhibit different patterns in the radius-angle space.

In the phase space, target tokens are generally located closer to the query direction with smaller angular distances, while background tokens tend to move toward larger angular distances, indicating that the angular component captures query-conditioned relevance. Meanwhile, the radius dimension provides additional structural cues by reflecting differences among target regions, temporal transitions, and background contexts.
The temporal response curves further illustrate this radius-angle behavior. Angular similarity shows stronger responses within the ground-truth intervals and lower responses in irrelevant regions, demonstrating its ability to identify query-aligned temporal evidence. Meanwhile, the radius curves exhibit distinct variations around target and transition regions, suggesting that the radial component captures complementary temporal structural information.

These observations demonstrate that the two geometric factors play complementary roles: angular direction models query-conditioned relevance, while radial distance provides structural cues for organizing temporal evidence. Together, they enable HTEO to construct a more interpretable hyperbolic representation space beyond flat Euclidean matching.

\subsection{Qualitative Analysis of Angular Relevance Refinement}
We further visualize how hyperbolic angular organization corrects false response peaks in representative cases. As shown in Fig.~\ref{fig:angle_hymr_examples}, we present four representative validation samples (QIDs 2343, 4625, 5926, and 10049), including cases where Euclidean similarity produces false peak responses outside the ground-truth moments.
For each sample, we compare the temporal response curves of Euclidean similarity and hyperbolic angular similarity. Euclidean similarity tends to assign stronger peak responses to background regions outside the target moments, resulting in distractor-induced false activations. After angular refinement, the dominant responses are shifted toward the ground-truth intervals, effectively suppressing these false peaks.
These examples demonstrate that angular organization provides a more reliable query-conditioned relevance measure, enabling HTEO to better distinguish target evidence from visually related but irrelevant temporal regions.

\subsection{Temporal Evidence Organization in Euclidean and Hyperbolic Spaces}

We visualize the organization of temporal tokens in Euclidean and hyperbolic spaces to examine whether HTEO provides a more structured representation of temporal evidence. As shown in Fig.~\ref{fig:token_distribution}, each point represents a temporal token projected into a 2D embedding space and is colored according to its grounding type, including target regions (GT), boundary transitions, and background contexts. We further report the silhouette score $S$ to quantify the separability of different evidence regions. 
The silhouette score ranges from $-1$ to $1$, with a higher value indicating greater intra-region compactness and clearer separation between different evidence regions.

In Euclidean space, temporal tokens with different grounding characteristics tend to overlap, making target evidence and background contexts difficult to distinguish. In contrast, hyperbolic geometry produces a more structured organization, where GT tokens form more coherent clusters, boundary tokens exhibit transitional patterns, and background tokens are better separated from query-relevant evidence. The consistently higher silhouette scores in hyperbolic space (e.g., $S=0.37$ vs. $S=0.28$) further demonstrate improved evidence separability.
These observations indicate that HTEO provides a more suitable geometric space for organizing heterogeneous temporal evidence, enabling clearer structural relationships among target moments, boundary transitions, and contextual backgrounds beyond flat Euclidean representations.

\subsection{Qualitative Analysis}

As shown in Fig.~\ref{fig:qualitative}, HTEO consistently generates more accurate temporal spans than TR-DETR and LD-DETR. In this figure, \textit{Ours} denotes HTEO built upon LD-DETR. While the baseline methods may include irrelevant context or miss portions of the target moment due to ambiguous video-text matching, HTEO better captures query-relevant evidence and produces predictions that are closer to the ground-truth boundaries. These examples demonstrate that hyperbolic evidence organization improves temporal reasoning by providing a more structured representation of relevant evidence and distracting temporal regions.

\section{Conclusion}
\label{sec:conclusion}
In this paper, we revisit video moment retrieval from the perspective of structured temporal evidence organization. Different from conventional flat video-text matching, we argue that a query-relevant  moment is supported by heterogeneous temporal evidence with different grounding characteristics, including target regions, boundary transitions, and contextual information. Based on this view, we propose HTEO, a hyperbolic temporal evidence organization framework that organizes query-aware temporal memory in a Poincar\'{e} ball before moment decoding. By exploiting radius-angle geometry, HTEO introduces complementary geometric cues into temporal memory: the radial coordinate provides a structural cue for organizing temporal evidence across different grounding characteristics, while the angular direction captures query-conditioned semantic relevance. The hyperbolically organized memory is then mapped back to the Euclidean tangent space and incorporated into the decoder through a residual memory update, preserving the original cross-modal representation while enhancing its geometric structure.
Extensive experiments and analyses demonstrate the effectiveness and interpretability of HTEO. The results show that hyperbolic geometry provides an effective mechanism for organizing temporal evidence and improving moment localization. This work highlights the potential of geometry-aware representation learning for more structured and interpretable video-language grounding.


 


\bibliographystyle{IEEEtran}
\bibliography{egbib}







\begin{IEEEbiography}[{\includegraphics[width=1in,height=1.25in,clip,keepaspectratio]{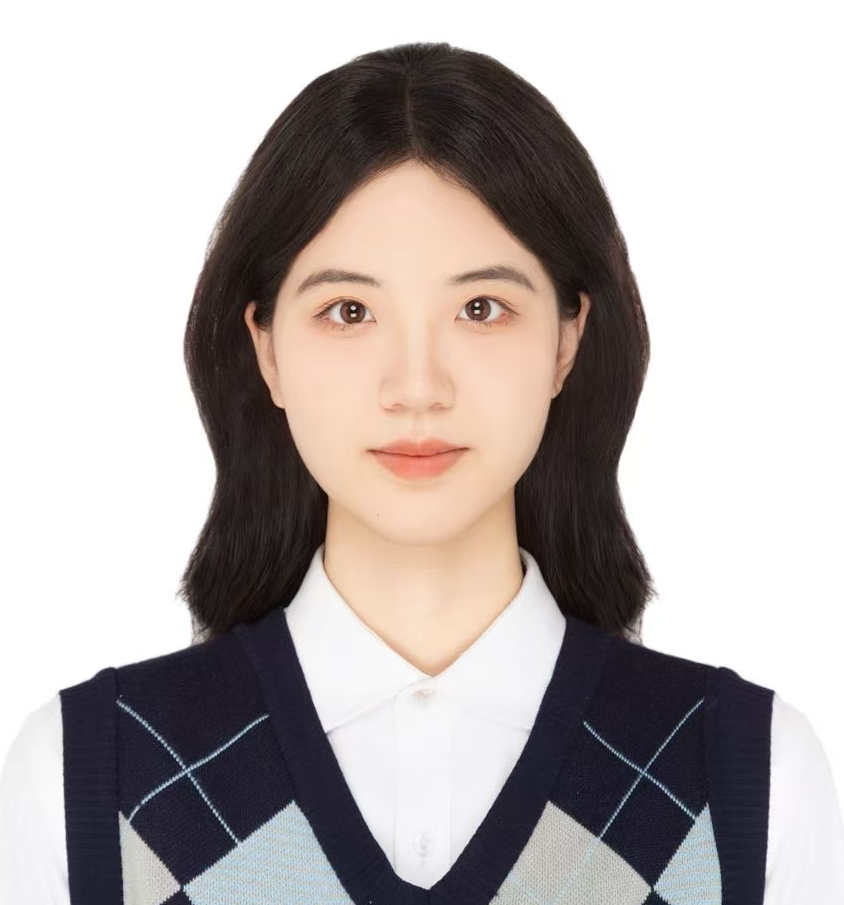}}]{Mengnan Liu} received the B.S. degree in Artificial Intelligence from Shandong University, Jinan, China, in 2022. She is currently pursuing the Ph.D. degree at the Institute of Artificial Intelligence and Robotics, Xi’an Jiaotong University. Her research interests include computer vision, multimodal video understanding and weakly-supervised learning.
\end{IEEEbiography}

\begin{IEEEbiography}[{\includegraphics[width=1in,height=1.25in,clip,keepaspectratio]{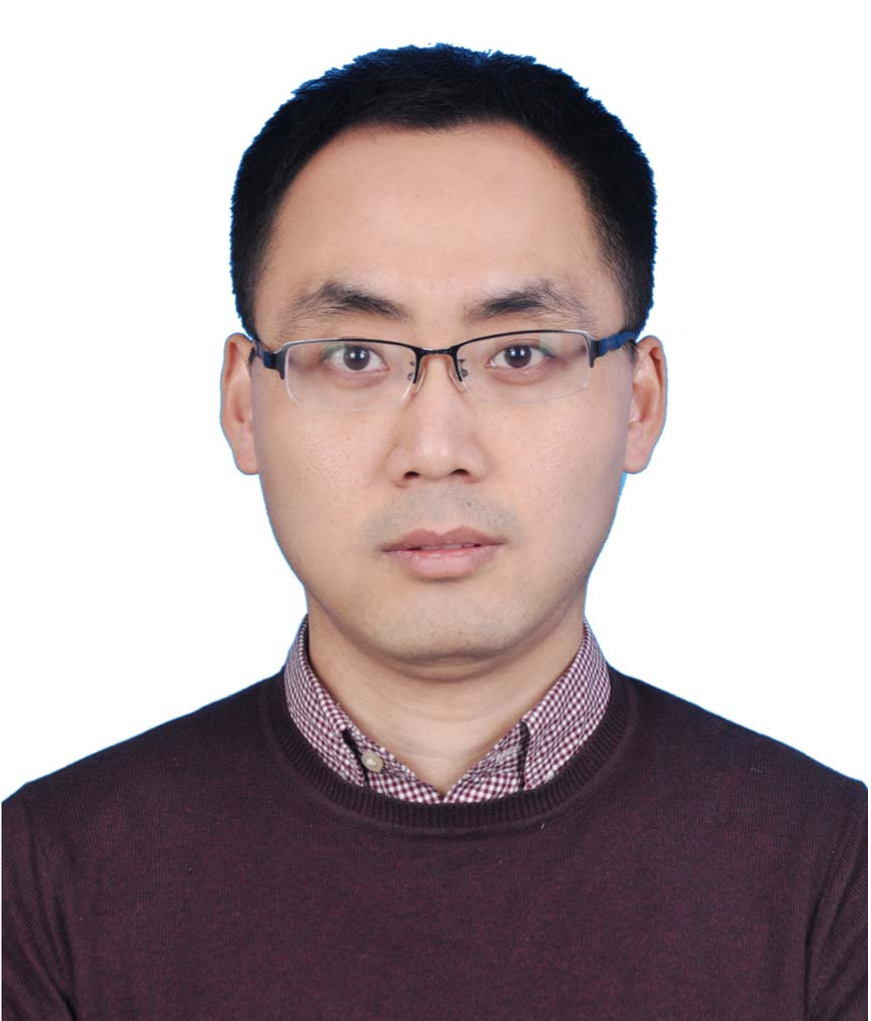}}]{Sanping Zhou}
(Member, IEEE) received the Ph.D. degree in control science and engineering from Xi'an Jiaotong University, Xi'an, China, in 2020. From 2018 to 2019, he was a Visiting Ph.D. Student with the Robotics Institute, Carnegie Mellon University. He is currently a Professor with the Institute of Artificial Intelligence and Robotics, Xi'an Jiaotong University. His research interests include machine learning, deep learning, and computer vision, with a focus on person re-identification, salient object detection, medical image segmentation, image classification, and visual tracking.
\end{IEEEbiography}

\begin{IEEEbiography}[{\includegraphics[width=1in,height=1.25in,clip,keepaspectratio]{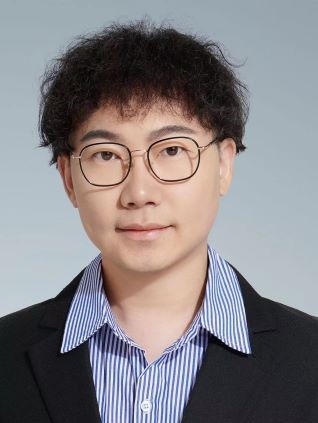}}]{Kun Xia}
received the Ph.D. degree in Control Science and Engineering from Xi'an Jiaotong University, Xi'an, China, in 2024. 
From 2022 to 2023, he was a visiting Ph.D. student with University of Illinois Chicago, IL, USA. He is currently an Assistant Professor with the School of Computer Science and Technology, Xi'an Jiaotong University, Xi'an, China. 
His research interests include computer vision, image/video processing, analysis and understanding.
\end{IEEEbiography}

\begin{IEEEbiography}
[{\includegraphics[width=1in,height=1.25in,clip,keepaspectratio]{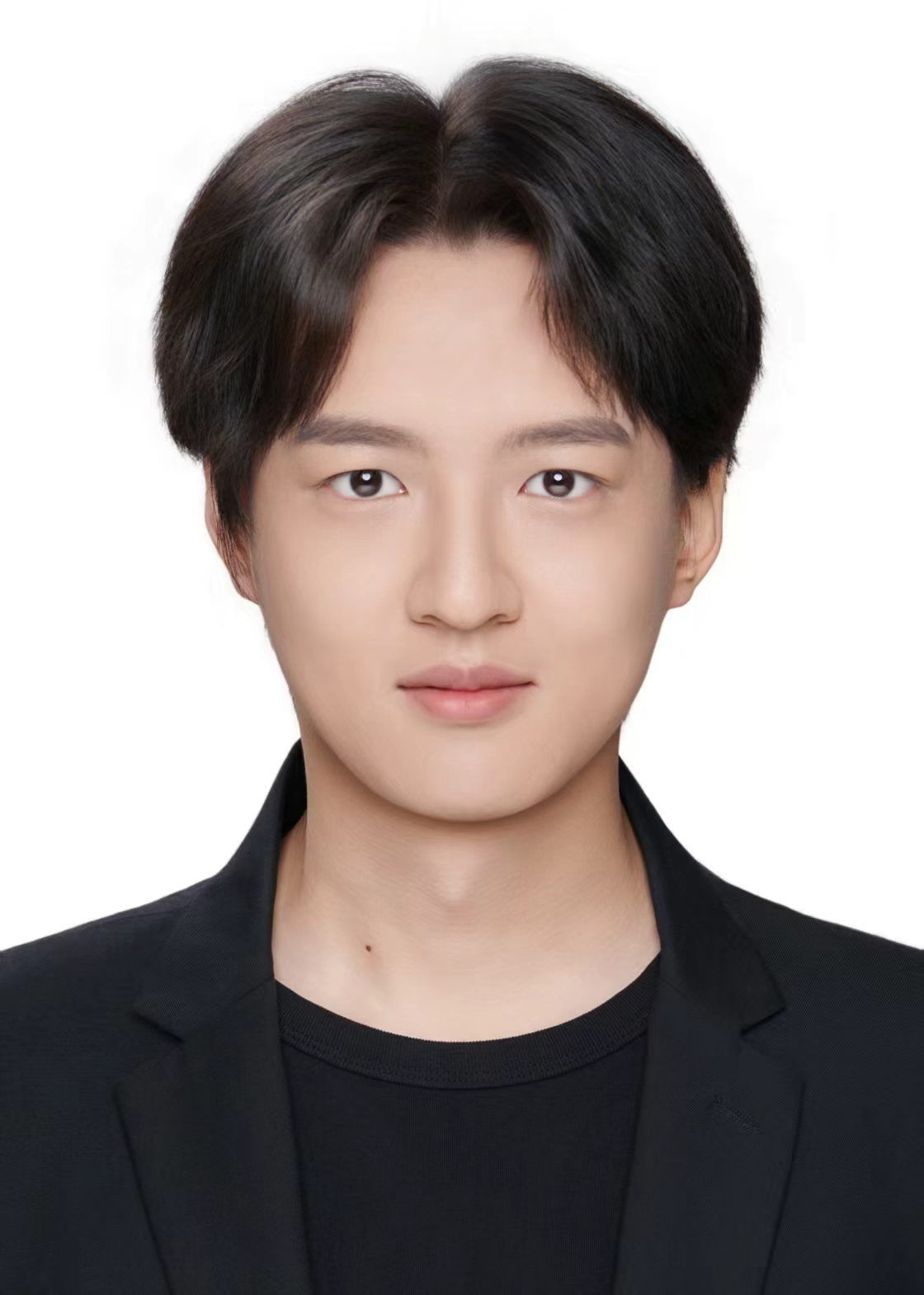}}]
{Jiaxuan Lin} received the B.S. degree in control
science and engineering in 2024 from Xi’an Jiaotong
University, Xi’an, China, where he is currently work-
ing toward the M.S. degree in artificial intelligence.
His research interests include computer vision, video
analysis and understanding.
\end{IEEEbiography}


\begin{IEEEbiography}[{\includegraphics[width=1in,height=1.25in,clip,keepaspectratio]{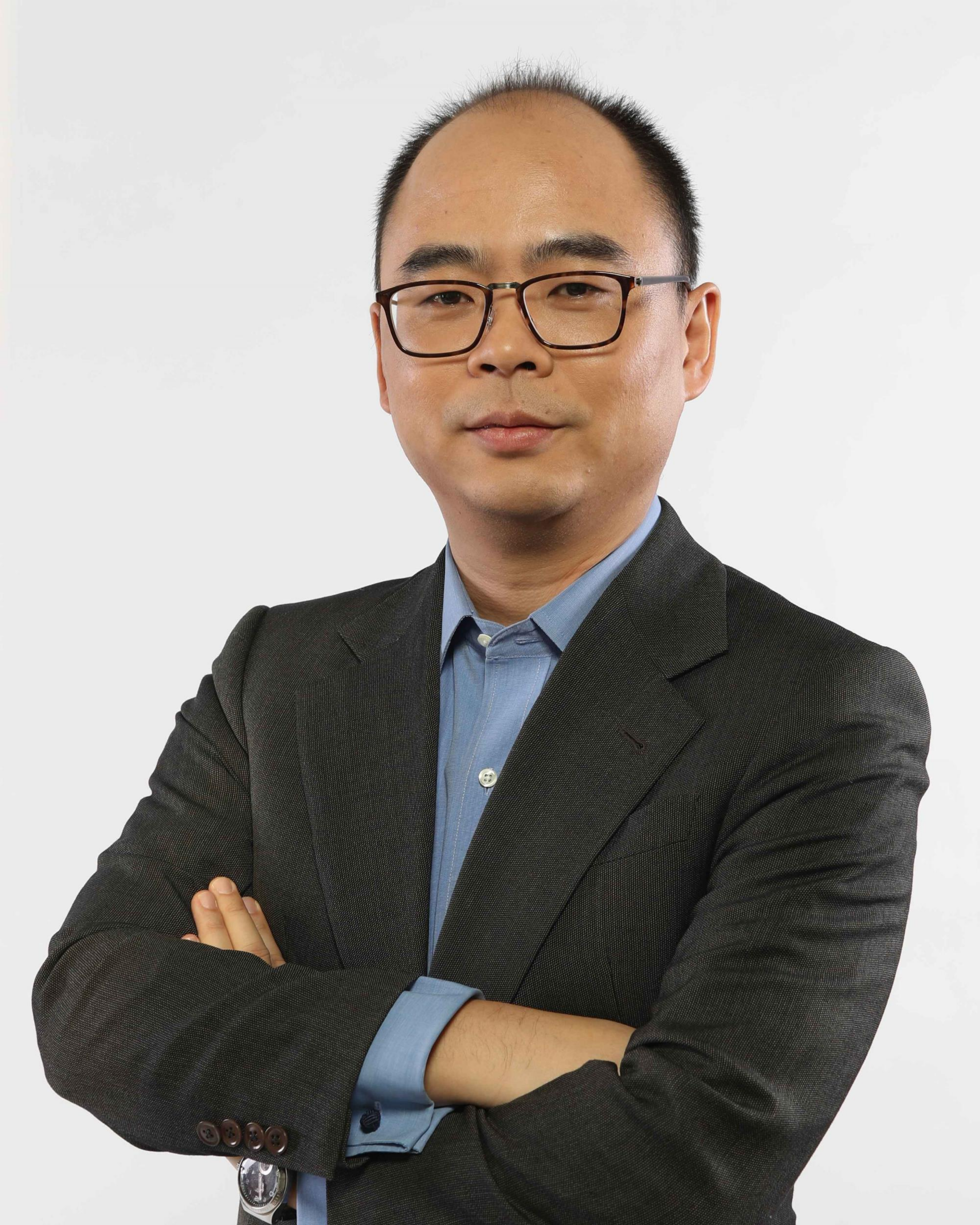}}]{Gang Hua}
(Fellow, IEEE) received the B.S. and M.S. degrees in Automatic Control Engineering from Xi’an Jiaotong University (XJTU), Xi’an, China, in 1999 and 2002, respectively. He received the Ph.D. degree in Electrical Engineering and Computer Science at Northwestern University, Evanston, Illinois, USA, in 2006. He is currently the  Director of Applied Science at Amazon Alexa AI. He is an associate editor of TPAMI and MVA. He is a general chair of ICCV’2025, a program chair of CVPR’2019\&2022, and an area chair of CVPR’2015\&2017 and ICCV’2011\&2017. He is the author of more than 200 peer reviewed publications in prestigious international journals and conferences. He holds 19 US patents and has 15 more US patents pending. He is the recipient of the 2015 IAPR Young Biometrics Investigator Award. He is an IEEE Fellow, an IAPR Fellow, and an ACM Distinguished Scientist.
\end{IEEEbiography}

\begin{IEEEbiography}[{\includegraphics[width=1in,height=1.25in,clip,keepaspectratio]{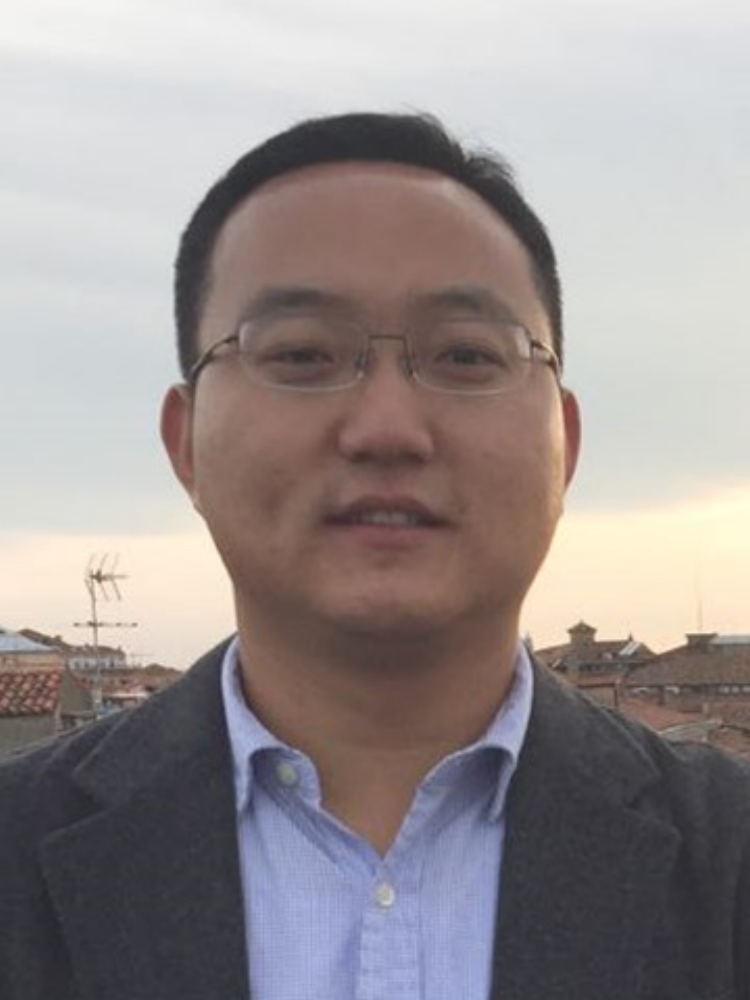}}]{Le Wang}
(Senior Member, IEEE) received the B.S. and Ph.D. degrees in Control Science and Engineering from Xi'an Jiaotong University, Xi'an, China, in 2008 and 2014, respectively. From 2013 to 2014, he was a visiting Ph.D. student with Stevens Institute of Technology, Hoboken, New Jersey, USA. From 2016 to 2017, he is a visiting scholar with Northwestern University, Evanston, Illinois, USA. He is currently a Professor with the Institute of Artificial Intelligence and Robotics of Xi'an Jiaotong University, Xi'an, China. His research interests include computer vision, pattern recognition, and machine learning. He is an associate editor of Pattern Recognition Letters. He is an area chair of CVPR'2022, ICME'2022 and IAPR'2022, and a senior program committee member of AAAI'2022. He holds 8 China patents and has 22 more China patents pending. He is the author of more than 60 peer reviewed publications in prestigious international journals and conferences.
\end{IEEEbiography}

\end{document}

%% file: tables-and-figures/fig-illustration.tex

\begin{figure*}[t]
  \centering
  \includegraphics[width=0.9\textwidth]{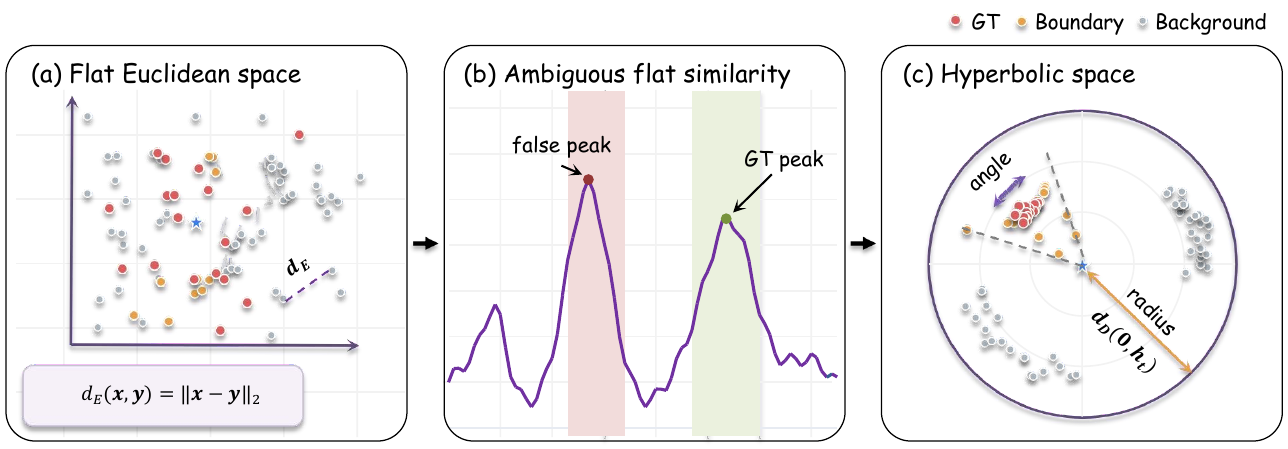}
  \caption{
Motivation of hyperbolic temporal evidence organization for video moment retrieval.
(a) In flat Euclidean memory, query-relevant tokens and contextual distractors may exhibit similar feature distances, making them difficult to distinguish based on flat similarity alone.
(b) Such ambiguity can lead to false response peaks on distracting background regions, resulting in inaccurate temporal localization.
(c) HTEO organizes query-aware temporal evidence in hyperbolic space through radius-angle geometry, where angular direction captures query-conditioned relevance and radial distance provides structural cues for temporal evidence organization. By introducing geometry-aware organization, HTEO enhances the discrimination between relevant evidence and semantically similar distractors, leading to more structured temporal representations for moment retrieval.}
\label{fig:illustration}

\end{figure*}

%% file: tables-and-figures/fig-framework.tex
\begin{figure*}[tb]
  \centering
  \includegraphics[width=\textwidth]{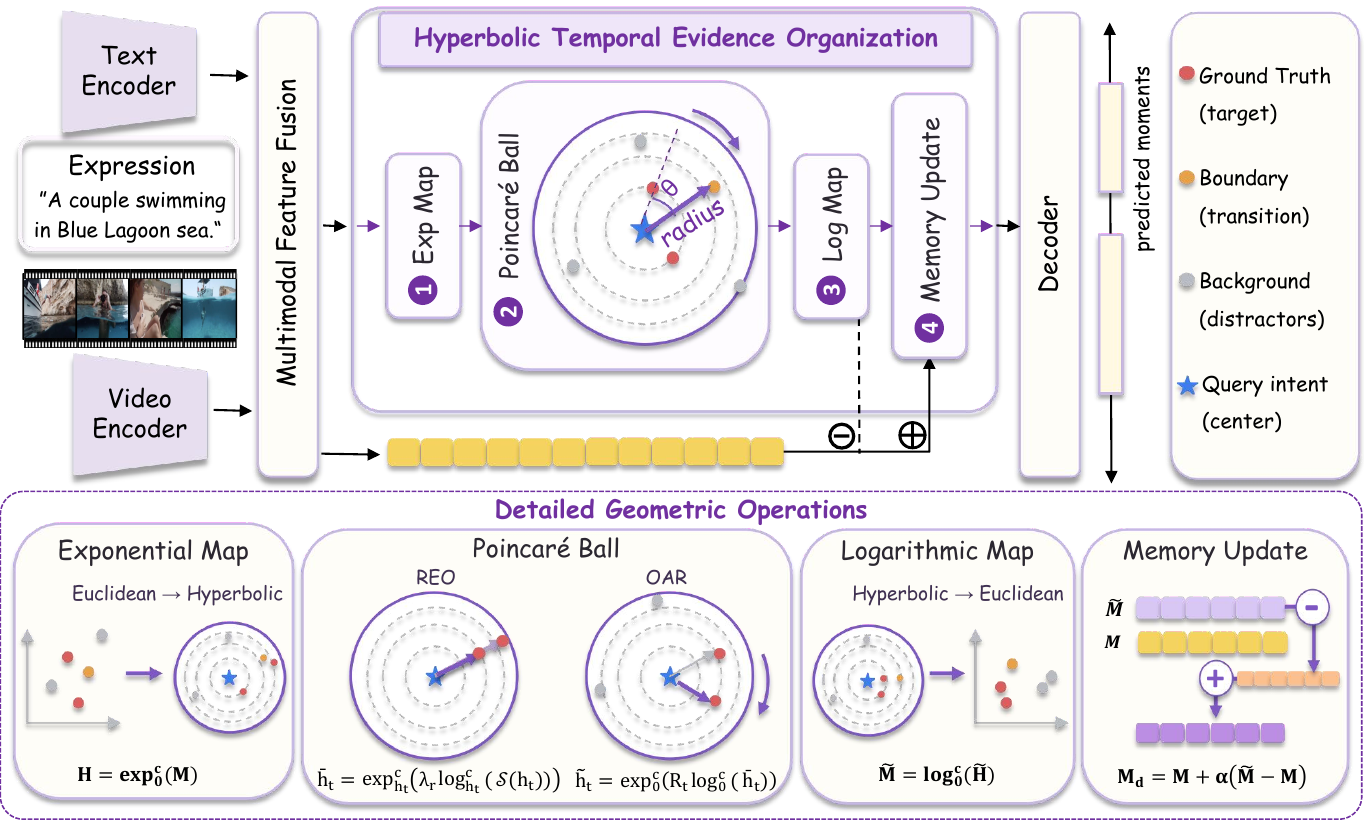}
\caption{Overview of the proposed Hyperbolic Temporal Evidence Organization (HTEO) framework.
Given a video-query pair, HTEO first constructs query-aware temporal memory through multimodal feature fusion and maps it into a Poincar\'{e} ball via the exponential map.
In hyperbolic space, temporal evidence is organized through Radial Evidence Organization (REO) and Orthogonal Angular Refinement (OAR), where REO regulates the radial structure to capture different temporal grounding characteristics, and OAR refines angular directions to enhance query-conditioned semantic relevance.
The organized memory is then projected back to the Euclidean tangent space via the logarithmic map and incorporated into the decoder through a lightweight residual update for moment localization.
The bottom panel illustrates the detailed geometric operations.}
	\label{fig:framework}

\end{figure*}

%% file: tables-and-figures/angle_speration.tex

\begin{figure*}[t]
    \centering
        \begin{minipage}[t]{0.515\textwidth}
        \centering
        \includegraphics[width=0.8\linewidth]{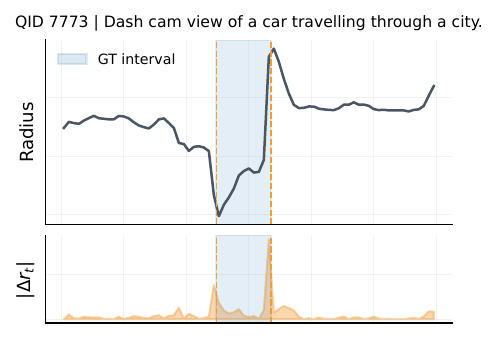}
        \par\small (a) Radius transition.
    \end{minipage}
    \hfill
    \begin{minipage}[t]{0.475\textwidth}
        \centering
        \includegraphics[width=0.79\linewidth]
{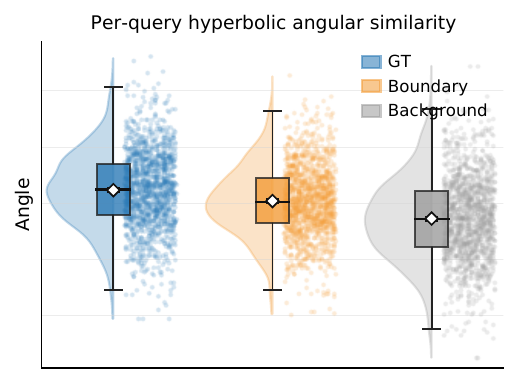}
        \par\small (b) Angular relevance.
    \end{minipage}
    \caption{
    Visualization of hyperbolic radius-angle behavior.}
    \label{fig:intro_radius_angle}
\end{figure*}

%% file: tables-and-figures/tab-comparison-on-thumos.tex
\begin{table*}[tb]
\centering
  \caption{Comparison of model performance on QVHighlights dataset. $^\dagger$ indicates the model using LLM.}
  \label{tab:qvhighligts}
\begin{tabular}{l c c c c c c c}
    \toprule
    \multirow{3}{*}{Method} & \multicolumn{5}{c}{MR} & \multicolumn{2}{c}{HD} \\  
    \cmidrule(lr){2-6}
    \cmidrule(lr){7-8}
    & \multicolumn{2}{c}{R1} & \multicolumn{3}{c}{mAP} & \multicolumn{2}{c}{$\geq$Very Good} \\ 
    \cmidrule(lr){2-3}
    \cmidrule(lr){4-6}
    \cmidrule(lr){7-8}
    & $@$0.5 & $@$0.7 & $@$0.5  & $@$0.75 & Avg. & mAP & HIT$@$1 \\
    \midrule
    M-DETR~\cite{lei2021detecting} (NeurIPS'2021) & 52.89 & 33.02 & 54.84 & 29.40 & 30.73 & 35.69 & 55.60 \\
    QD-DETR~\cite{moon2023query} (CVPR'2023) & 62.52 & 44.98 & 62.62 & 39.88 & 39.86 & 38.64 & 62.40 \\
    UniVTG~\cite{lin2023univtg} (ICCV'2023) & 66.65 & 52.19 & 64.37 & 46.68 & 45.18 & 40.18 & 64.77  \\
    TaskWeave~\cite{yang2024task} (CVPR'2024) & 64.26 & 50.06 & 65.39 & 46.47 & 45.38 & - & - \\
   \rowcolor{green!10}  TR-DETR~\cite{sun2024tr} (AAAI'2024) & 67.10 & 51.48 & 66.27 & 45.09 & 46.42 & 40.55 & 64.77 \\
    LLMEPET$^\dagger$~\cite{jiang2024prior} (ACM MM'2024) & 66.58 & 51.10 & - & - & 46.24 & - & - \\
        LMR$^\dagger$~\cite{liu2025context} (TMM'2025) & 64.40 & 47.21 & 64.65 & 43.16 & 42.56 & - & - \\
     \rowcolor{cyan!10}LD-DETR~\cite{zhao2025ld} (ICCV'2025) &\underline{67.87}& 51.35 &\underline{67.23}&\underline{47.29}&\underline{47.18}& \underline{41.39} &\underline{66.39}\\
 \midrule
\rowcolor{green!10} TR-DETR + HTEO (Ours) & 67.55 & 51.86 & 66.98 & 45.68 & 47.63 & 40.99 & 65.29 \\
    \rowcolor{cyan!10}
LD-DETR + HTEO (Ours)& \textbf{68.65} & \textbf{53.74} & \textbf{68.11} & \textbf{49.34} & \textbf{48.42} & \textbf{42.08} & \textbf{68.26} \\
    \bottomrule
  \end{tabular}
\end{table*}

%% file: tables-and-figures/tab-comparison-on-activitynet13.tex
\begin{table*}[t]
\centering
\caption{Comparison of model performance on TACoS and Charades-STA datasets.}

  \centering
  \small 
  \begin{tabular}{l c c c c c c c c}
    \toprule
   \multirow{2}{*}{Method} & \multicolumn{4}{c}{TACoS} & \multicolumn{4}{c}{Charades-STA} \\  
    \cmidrule(r){2-5}
    \cmidrule(r){6-9}
 & R1$@$0.3 & R1$@$0.5 & R1$@$0.7 & mIoU &  R1$@$0.3 & R1$@$0.5 & R1$@$0.7 & mIoU\\
\midrule
M-DETR~\cite{lei2021detecting} (NeurIPS'2021) & 38.0 & 24.7 & 12.0 & 25.5 & 65.8 & 52.1 & 30.6 & 45.5  \\
MomentDiff~\cite{li2023momentdiff} (NeurIPS'2023) & 44.8 & 33.7 & - & - & - & 55.6 & 32.4 & -  \\
UniVTG~\cite{lin2023univtg} (ICCV'2023) & - & - & - & - & 68.0 & 57.2 & 36.1 & - \\
CG-DETR~\cite{moon2023correlation} (arXiv’2023)  & 52.2 & 39.6 & 22.2 & 36.5 & 70.4 & 58.4 & 36.3 & 50.1 \\
TR-DETR~\cite{sun2024tr} (AAAI'2024) & - & - & - & - & - & 57.6 & 33.5 & -  \\
LLMEPET$^\dagger$~\cite{jiang2024prior} (ACM MM'2024) & 52.7 & - & 22.8 & 36.6 & 70.9 & - & 36.5 & 50.3 \\
LMR$^\dagger$~\cite{liu2025context} (TMM'2025)& - & - & - & - & - & 55.9 & 35.2 & - \\
LD-DETR~\cite{zhao2025ld} (ICCV'2025) & \underline{57.6} & \underline{44.3} & \underline{26.2} & \underline{40.3} & \underline{73.9} & \underline{62.6} & \underline{41.6} & \underline{53.4} \\
\rowcolor{cyan!10}
LD-DETR + HTEO &  \textbf{58.4} & \textbf{45.2} & \textbf{27.0} & \textbf{41.5} & \textbf{74.5} & \textbf{63.6} & \textbf{42.9} &  \textbf{54.1}  \\

    \bottomrule
  \end{tabular}

   \label{tab:tacos and charades}
\end{table*}

%% file: tables-and-figures/fig-hyperbolic.tex
\begin{figure*}[t]
    \centering
    \begin{minipage}[t]{0.495\textwidth}
        \centering
        \includegraphics[width=\linewidth]{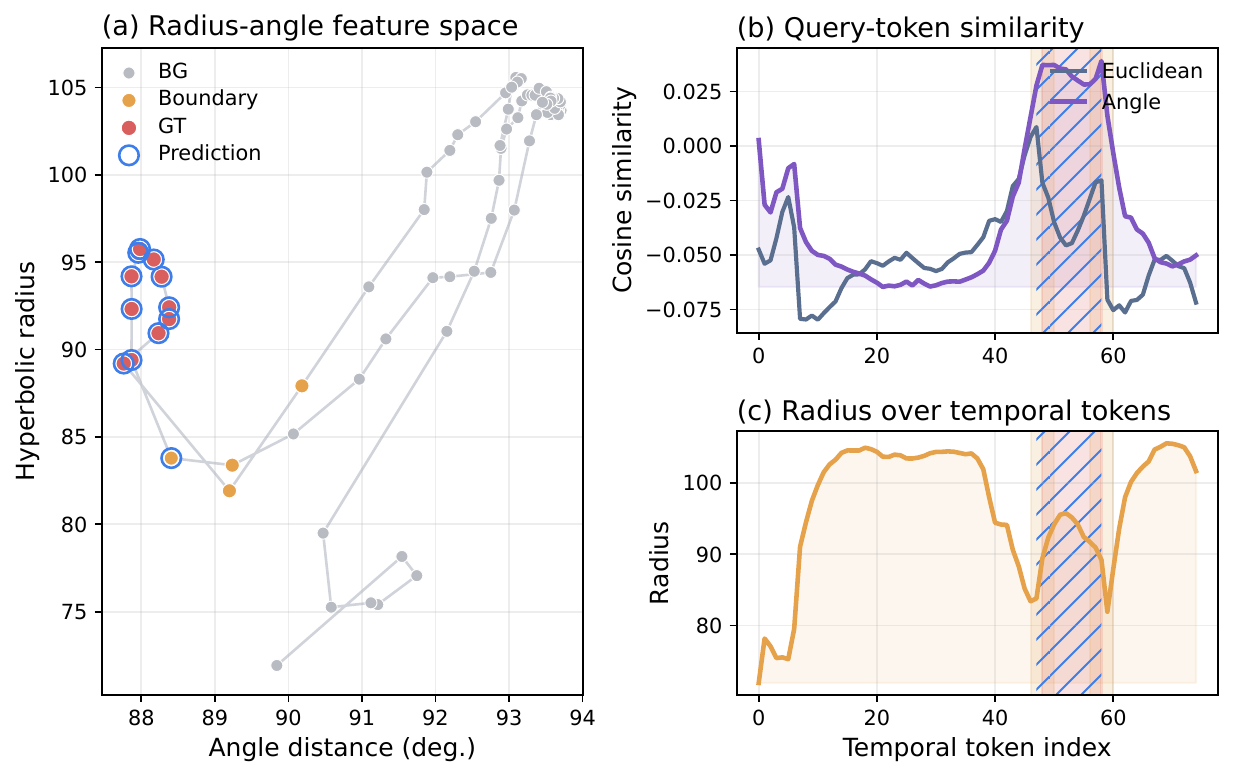}
        \par\small
    \end{minipage}\hfill%
    \begin{minipage}[t]{0.495\textwidth}
        \centering
        \includegraphics[width=\linewidth]{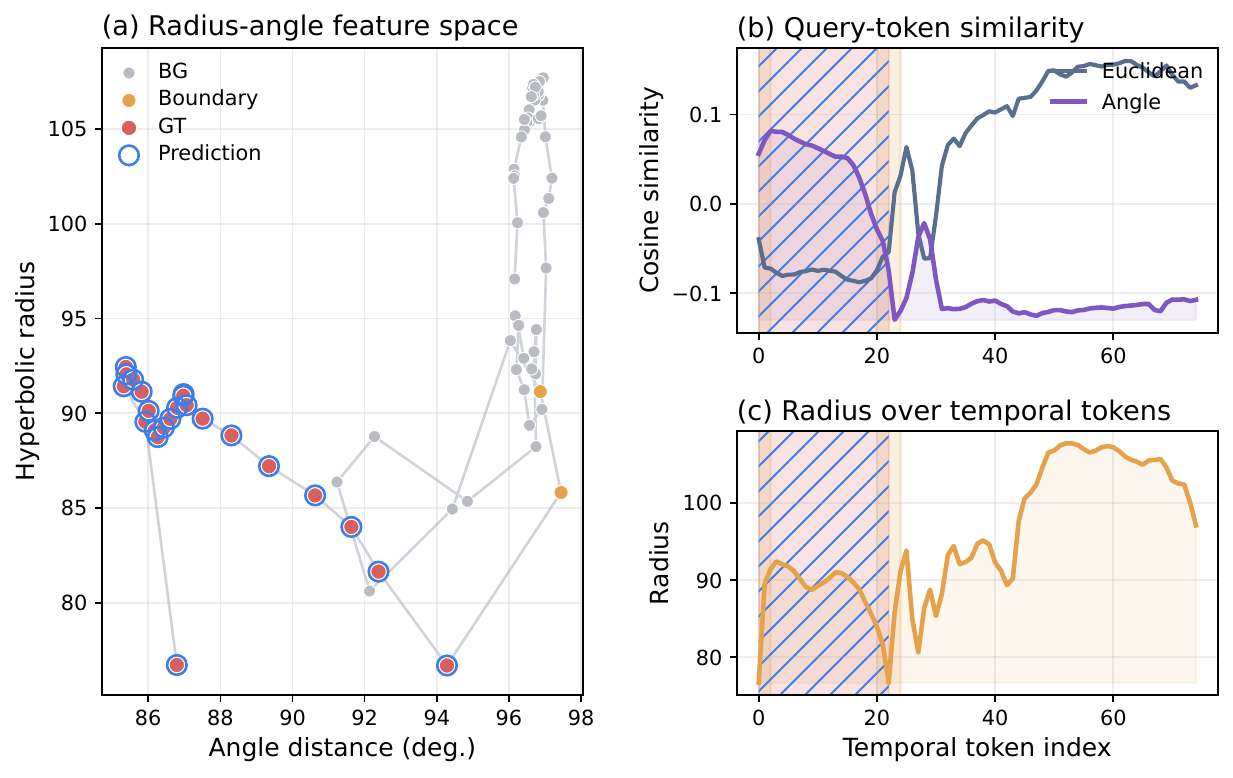}
        \par\small 
    \end{minipage}
\caption{
Visualization of hyperbolic radius-angle organization for temporal evidence modeling.
(a) The radius-angle feature space shows the geometric separation among target, boundary, and background tokens.
(b) The angular similarity curves demonstrate that angular direction provides query-conditioned relevance cues.
(c) The temporal evolution of radius reveals structural variations around different temporal evidence regions and boundary transitions.
}
    \label{fig:radius_angle}
\end{figure*}

%% file: tables-and-figures/tab-ablation-radius-angle.tex
  \begin{table}[h]
  \centering
  \caption{Ablation study on the contributions of radius and angle components in HTEO.}
  \label{tab:radius-angle calibration}
   \small
  \begin{tabular}{ccccc}
  \toprule
  Radius & Angle & R1@0.5 &R1@0.7 & Avg. mAP \\
  \midrule
  \xmark & \xmark &67.87 & 51.35 & 47.18 \\
  \cmark & \xmark & 69.55 & 54.39 & 47.98\\
 \xmark & \cmark & 68.06 & 52.58 & 48.21 \\
 \cmark & \cmark & 68.65 & 53.74 & 48.42 \\
  \bottomrule
  \end{tabular}
  
  \end{table}

%% file: tables-and-figures/tab-ablation-loss.tex

\begin{table}[h]
\centering
\caption{Ablation study on geometric supervision for hyperbolic temporal evidence organization.}
\label{tab:geometric losses}
\small
\setlength{\tabcolsep}{4pt}

\begin{tabular}{cc|ccc}
\toprule
$\lambda_{rad}$ & $\lambda_{ang}$ & R1@0.5 & R1@0.7 & Avg. mAP \\
\midrule
 0.00 & 0.00 &67.87 & 51.35 & 47.18 \\
\hline
0.25 & 0.00 & 67.55 & 52.90 & 47.88 \\
0.50 & 0.00 & 69.55 & 54.39 & 47.98\\
1.00 & 0.00 & 67.48 & 53.81 & 47.94 \\
\hline
0.00 & 0.25 & 68.06 & 53.23 & 47.75 \\
0.00 & 0.50& 68.06 & 52.58 & 48.21 \\
0.00 & 1.00 & 68.65 & 53.03 & 47.59  \\
\hline
0.50 & 0.50 & 68.65 & 53.74 & 48.42 \\
\bottomrule
\end{tabular}
\end{table}

%% file: tables-and-figures/tab-background-confusion.tex
\begin{table}[h]
\centering
 \small
\setlength{\tabcolsep}{4pt}
\caption{Robustness analysis under different levels of temporal evidence ambiguity.}
\label{tab:Background confusion analysis}
\begin{tabular}{lcccc}
\toprule
Confusion & Method& R1@0.5 & R1@0.7 & Avg. mAP\\
\midrule
Low & Baseline& 76.95 & 69.14 & 73.52\\
Low & HTEO & 82.36 & 73.75 & 78.81 \\
Medium & Baseline & 53.20 & 39.80 & 48.85\\
Medium & HTEO  & 53.80 & 39.60 & 49.13   \\
High & Baseline & 35.03 & 24.32 & 34.94  \\
High & HTEO  & 40.47 & 28.68 & 38.80 \\
\bottomrule
\end{tabular}
\end{table}

%% file: tables-and-figures/tab-moment-length.tex
\begin{table}[h]
\centering
 \small
\setlength{\tabcolsep}{4pt}
\caption{Analysis of HTEO under different moment durations.}
\label{tab:momoent length}
\begin{tabular}{lcccc}
\toprule
Length & Method  & R1@0.5 & R1@0.7  & Avg. mAP \\
\midrule
Short & Baseline  & 12.59 & 6.76 & 11.82 \\
Short & HTEO & 14.22 & 9.09 & 13.34\\
Medium & Baseline & 65.73 & 51.02& 49.89 \\
Medium & HTEO & 66.25 &51.31  & 50.31 \\
Long & Baseline  &64.29 & 52.28  & 53.50 \\
Long & HTEO &65.85 & 52.79  & 53.92 \\
\bottomrule
\end{tabular}
\end{table}

%% file: tables-and-figures/peak.tex
\begin{figure}[h]
  \centering
  \includegraphics[width=\linewidth]{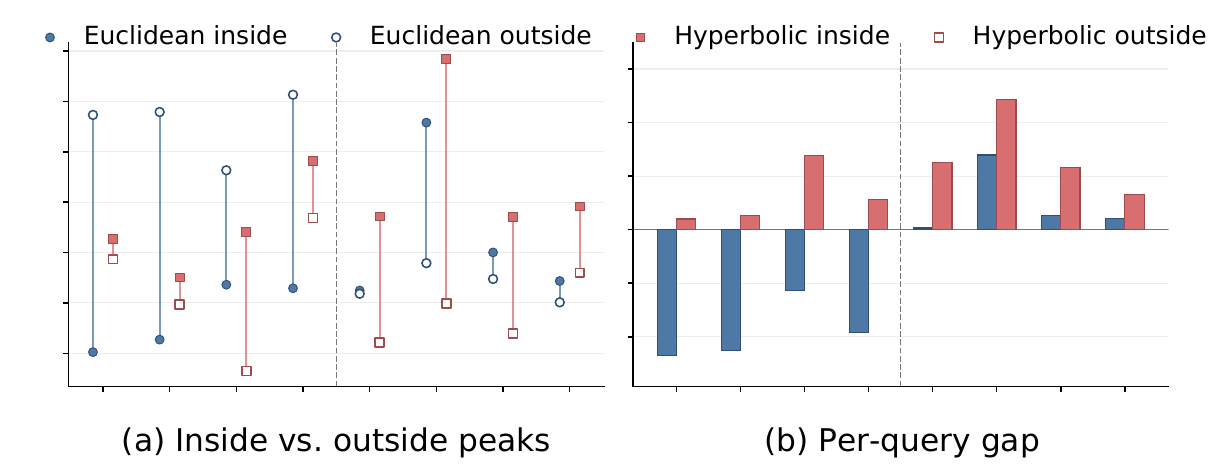}
\caption{Illustrative analysis of inside-outside response gaps.
The left panel compares the maximum query-conditioned responses inside and outside the ground-truth interval, while the right panel shows the corresponding per-query inside-outside response gaps. }
	\label{fig:false_peak}

\end{figure}

%% file: tables-and-figures/fig-angle.tex
\begin{figure*}[t]
\centering
\begin{minipage}[t]{0.495\linewidth}
  \centering
  \includegraphics[width=\linewidth]{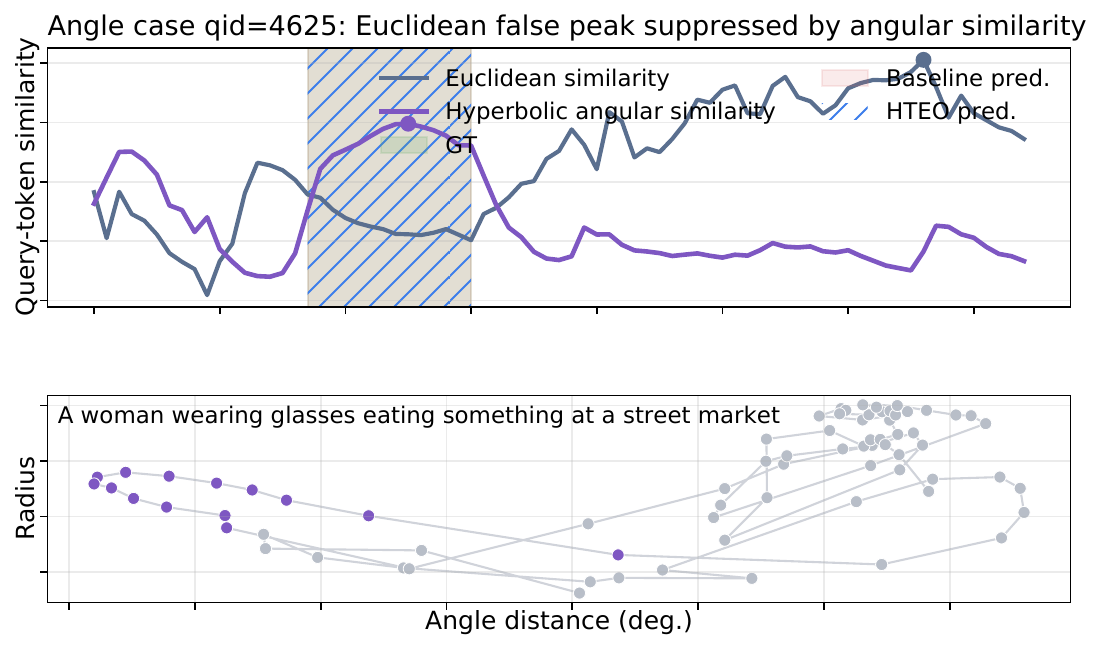}
 
\end{minipage}
\hfill
\begin{minipage}[t]{0.495\linewidth}
  \centering
  \includegraphics[width=\linewidth]{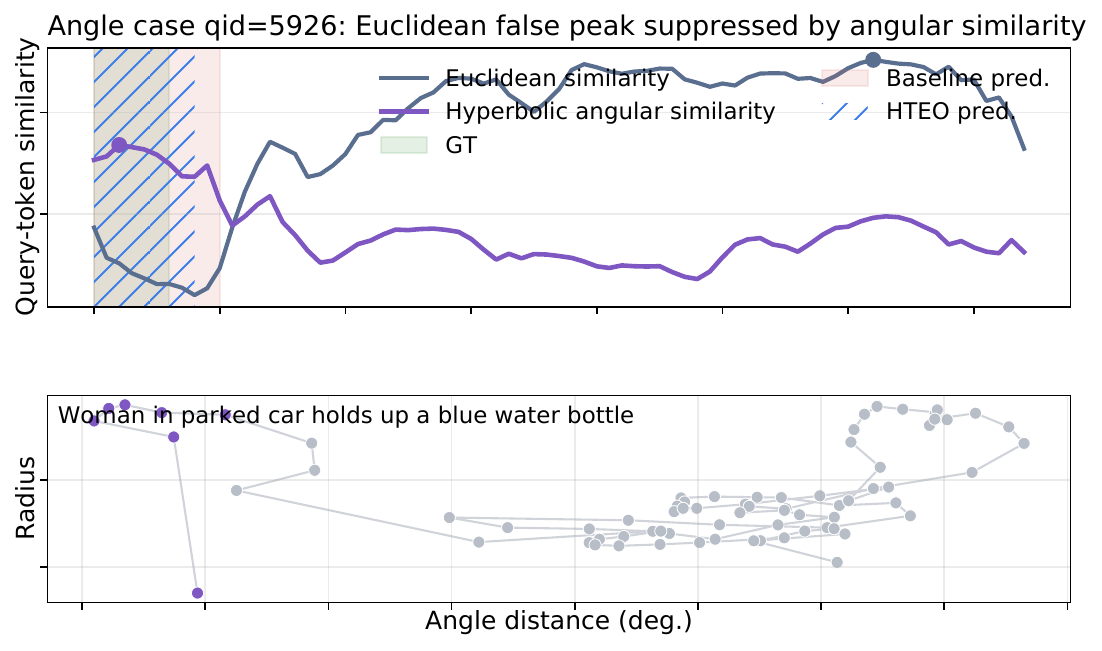}
 
\end{minipage}

\vskip 0.3cm 

\begin{minipage}[t]{0.495\linewidth}
  \centering
  \includegraphics[width=\linewidth]{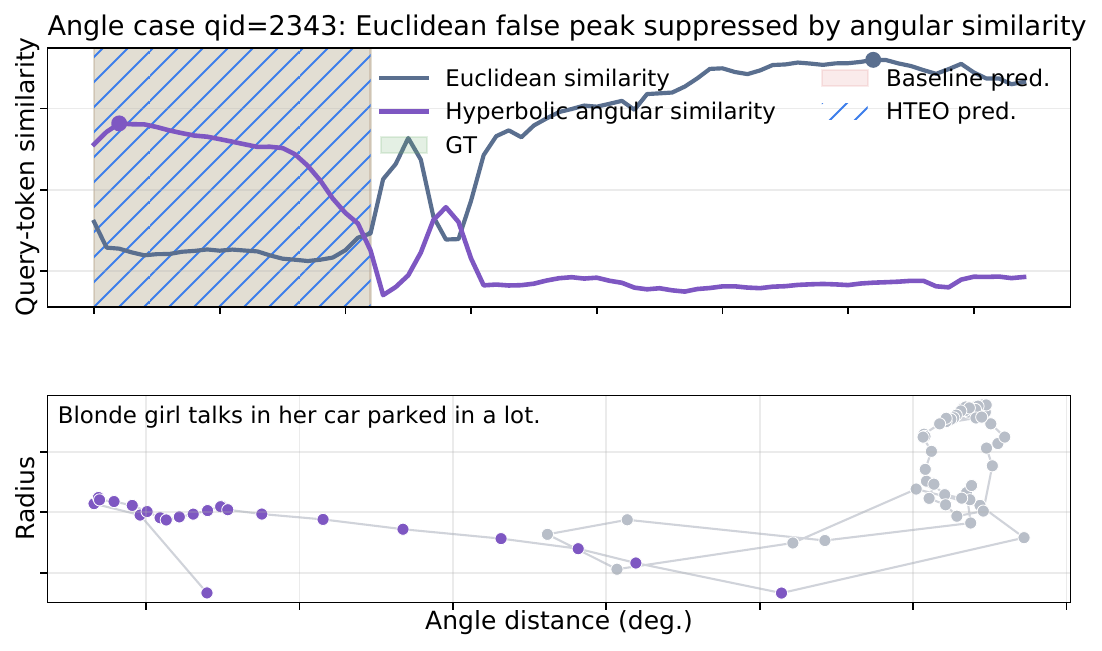}
 
\end{minipage}
\hfill
\begin{minipage}[t]{0.495\linewidth}
  \centering
  \includegraphics[width=\linewidth]{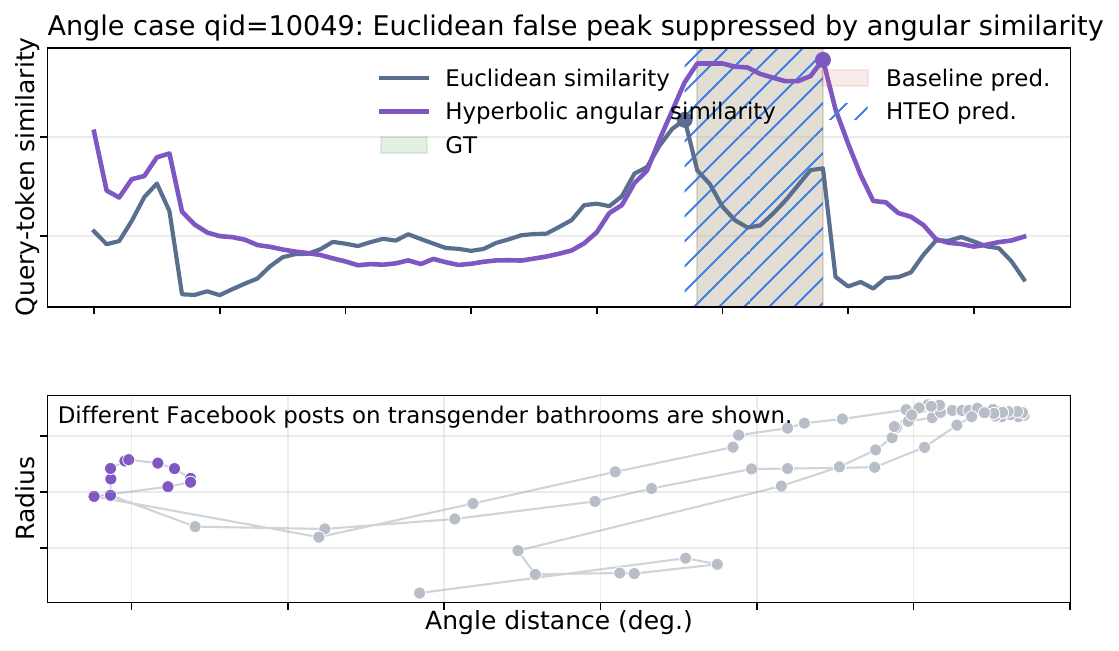}

\end{minipage}

\caption{Qualitative analysis of false peak correction using hyperbolic angular refinement. 
Four representative validation samples are shown where Euclidean similarity produces distractor-induced response peaks. Hyperbolic angular organization shifts dominant responses toward the ground-truth moments and suppresses irrelevant activations.}

\label{fig:angle_hymr_examples}
\end{figure*}

%% file: tables-and-figures/tab-radius_structure.tex
\begin{table}[h]
\centering
\small
\setlength{\tabcolsep}{4pt}
\caption{Analysis of radial statistics in hyperbolic space. 
Mean, median, and standard deviation describe the radius distributions of different temporal regions, while Dist. denotes the geodesic distance to the GT moment center.}
\label{tab:radius_structural_analysis}
\begin{tabular}{lcccc}
\toprule
Region & Mean & Med. & Std. & Dist. (mean $\pm$ std) \\
\midrule
Boundary      & 29.14 & 29.11 & 0.5012 & 0.5932 $\pm$ 0.0673 \\
GT  & 28.75 & 28.77 & 0.3969 & 0.2490 $\pm$ 0.1442 \\
Background   & 29.51 & 29.65 & 0.4879 & 2.8878 $\pm$ 2.2725 \\

\bottomrule
\end{tabular}
\end{table}

%% file: tables-and-figures/tab-ablation-control.tex
\begin{table}[h]
\centering
 \small
\setlength{\tabcolsep}{4pt}
\caption{Ablation study of the residual memory update.}
\label{tab:euclidean_adapter_control}
\begin{tabular}{lccc}
\toprule
Method & R1@0.5 & R1@0.7 & Avg. mAP \\
\midrule
Baseline &67.87 & 51.35 & 47.18 \\
Euclidean MLP Adapter & 68.06 & 52.97 & 47.30 \\
Euclidean Residual Adapter & 67.48 & 52.19 & 47.32 \\
HTEO & {68.65} & {53.74} & {48.42} \\
\bottomrule
\end{tabular}
\end{table}

%% file: tables-and-figures/tab-ablation-curvaturer.tex
\begin{table}[h]
\centering
 \small
\setlength{\tabcolsep}{4pt}
\caption{Sensitivity to the curvature magnitude \(c\).}
\label{tab:Curvature sensitivity}
\begin{tabular}{lcccc}
\toprule
$c$ & R1@0.5 & R1@0.7 & Avg. mAP & mIoU  \\
\midrule
0.001 &   67.03&  53.29&  47.31 & 62.31 \\
0.005 & 67.74 & 52.65 & 47.63 &  62.83\\
0.010 & 68.65 & 53.74 & 48.42 & 63.90 \\
0.050  & 67.55 & 52.77 & 47.57 &  62.90\\
0.100  & 67.16 & 52.45 & 46.75 & 62.43 \\
\bottomrule
\end{tabular}
\end{table}

%% file: tables-and-figures/token_distribution.tex



\begin{figure*}[t]
    \centering

    \begin{minipage}[t]{0.495\textwidth}
        \centering
        \includegraphics[width=\linewidth]{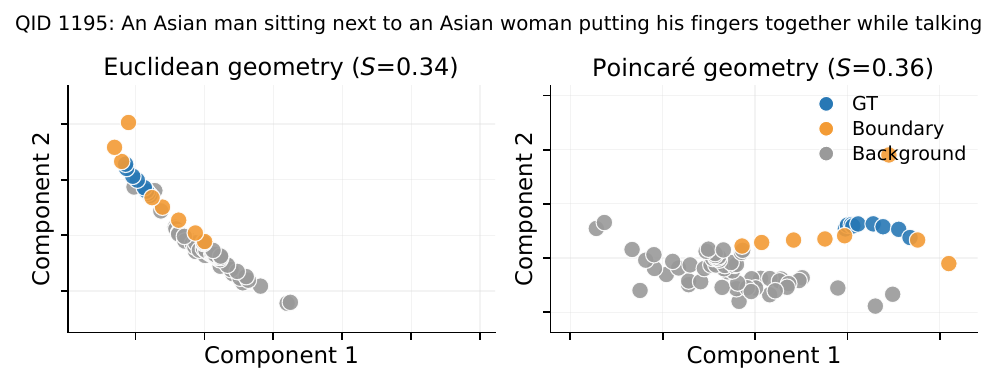}
        \par\small 
    \end{minipage}
    \hfill
    \begin{minipage}[t]{0.495\textwidth}
        \centering
        \includegraphics[width=\linewidth]{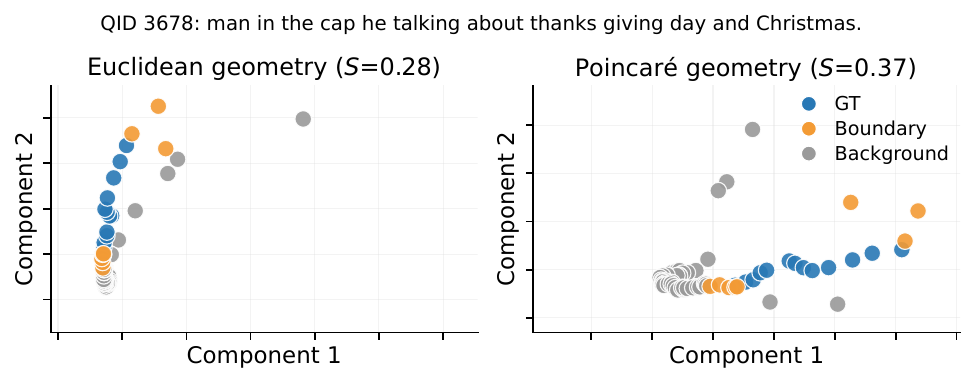}
        \par\small 
    \end{minipage}

    \vspace{0.8em}

    \begin{minipage}[t]{0.495\textwidth}
        \centering
        \includegraphics[width=\linewidth]{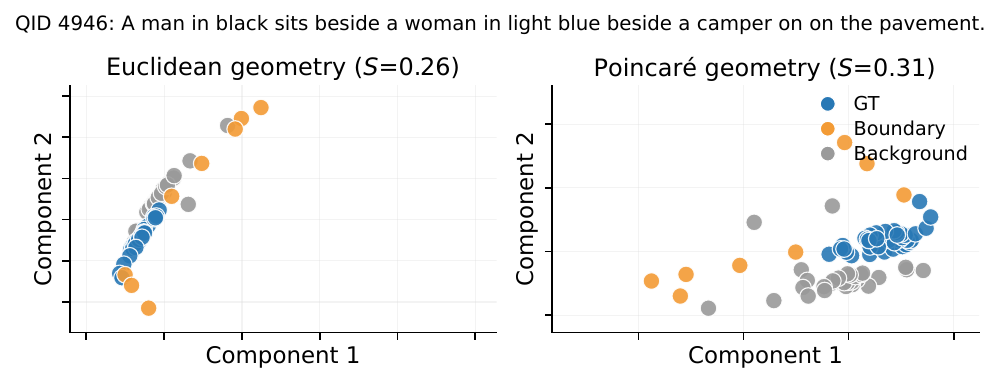}
        \par\small
    \end{minipage}
    \hfill
    \begin{minipage}[t]{0.495\textwidth}
        \centering
        \includegraphics[width=\linewidth]{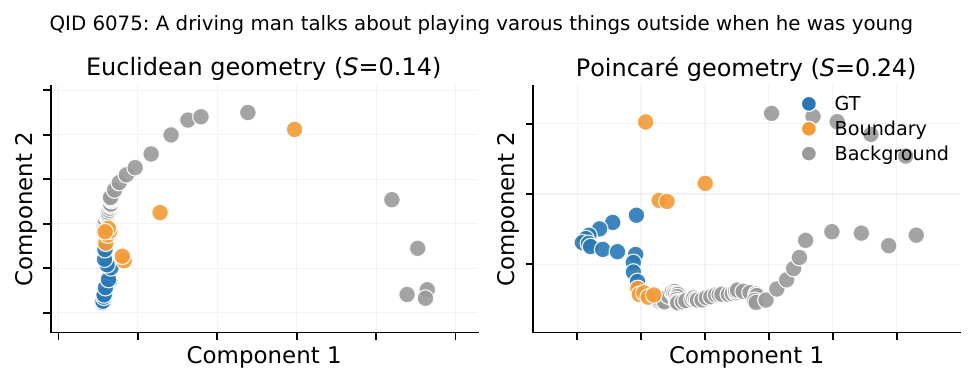}
        \par\small
    \end{minipage}

   \caption{Visualization of temporal evidence organization in Euclidean and hyperbolic spaces. 
Compared with flat Euclidean representations, hyperbolic geometry yields more structured distributions of target, boundary, and background tokens, indicating improved evidence separability. The query texts are displayed as provided in the original dataset.}
    \label{fig:token_distribution}
\end{figure*}

%% file: tables-and-figures/tab-position-cal.tex
\begin{table}[h]
\centering
 \small
\setlength{\tabcolsep}{4pt}
\caption{Ablation study on the injection position of HTEO.}
\label{tab:hyperbolic calibration}
\begin{tabular}{lccc}
\toprule
Position & R1@0.5 & R1@0.7 & Avg. mAP  \\
\midrule
None & 67.87   & 51.35 & 47.18\\
Text-only  &68.19  & 53.03 & 47.62  \\
Video-only & 68.13 & 53.16 &  47.92 \\
Pre-fusion & 68.45 & 53.55 & 47.39  \\
Post-fusion & 68.65 & 53.74   & 48.42\\
\bottomrule
\end{tabular}
\end{table}

%% file: tables-and-figures/tab-ablation-residual.tex
\begin{table}[h]
\centering
 \small
\setlength{\tabcolsep}{4pt}
\caption{Sensitivity to the residual scaling factor \(\alpha\).}
\label{tab:Delta scale alpha sensitivity}
\begin{tabular}{lcccc}
\toprule
$\alpha$ & R1@0.5 & R1@0.7 & Avg. mAP & mIoU \\
\midrule
0.01 & 67.35 & 52.84 & 47.88 & 63.16\\
0.05 & 67.74 & 53.35 & 47.80 & 63.20 \\
0.10 & 68.65 & 53.74 & 48.42 & 63.90\\
0.15 & 67.29 & 52.71 & 47.80 & 62.86\\
0.20 & 67.29 & 52.00 & 47.46 & 62.98  \\
\bottomrule
\end{tabular}
\end{table}

%% file: tables-and-figures/fig-qualitative.tex
\begin{figure*}[t]
  \centering
  \includegraphics[width=\textwidth]{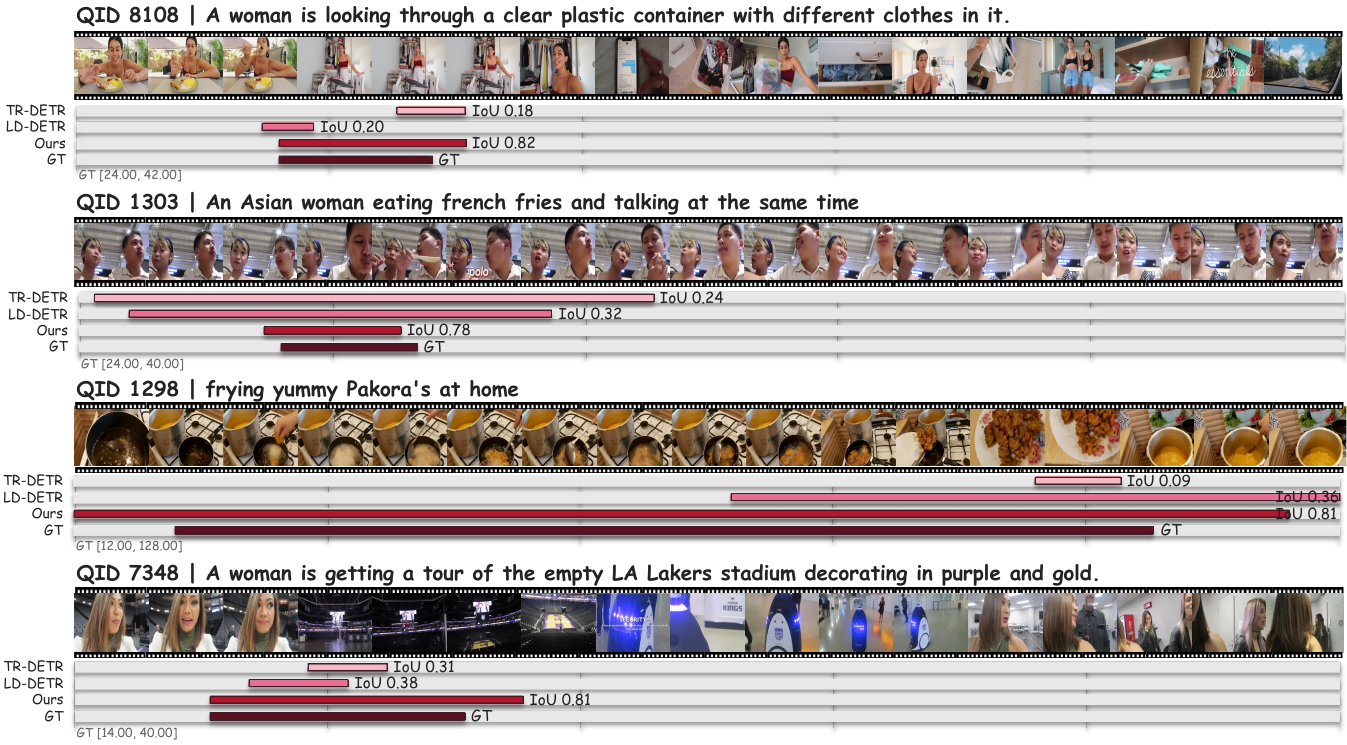}
 \caption{Qualitative comparison of HTEO with TR-DETR and LD-DETR on the QVHighlights dataset. HTEO achieves more accurate temporal localization with better alignment to GT. The queries are displayed as provided in the original dataset.}
\label{fig:qualitative}

\end{figure*}